\documentclass[sigconf]{acmart}
\AtBeginDocument{%
  \providecommand\BibTeX{{%
    \normalfont B\kern-0.5em{\scshape i\kern-0.25em b}\kern-0.8em\TeX}}}

\def\ie{{\em i.e.}}

\def\etal{{\em et al.}}

\newcommand{\figref}[1]{Fig. \ref{#1}}
\newcommand{\tabref}[1]{Tab. \ref{#1}}

\usepackage{multirow}
\usepackage{booktabs}
\usepackage{enumerate}
\usepackage{hyperref}
\hypersetup{colorlinks=true,linkcolor=red,urlcolor=red}

\begin{document}

\title{Partitioned Saliency Ranking with Dense Pyramid Transformers}

\author{Chengxiao Sun}
\authornote{Both authors contributed equally to this research.}

\affiliation{School of Software Engineering, 
  \institution{Huazhong University of Science and Technology}
  \city{Wuhan}
  \country{China}
}
\email{sunchengxiao@hust.edu.cn}

\author{Yan Xu}
\authornotemark[1]
 \affiliation{School of Software Engineering, 
   \institution{Huazhong University of Science and Technology}
  \city{Wuhan}
  \country{China}
}
\email{yan_xu@hust.edu.cn}

\author{Jialun Pei}
 \affiliation{School of Computer Science and Engineering, 
   \institution{The Chinese University of Hong Kong}
  \state{Hong Kong}
  \country{China}
}
\email{jialunpei@cuhk.edu.hk}

\author{Haopeng Fang}
 \affiliation{School of Software Engineering, 
   \institution{Huazhong University of Science and Technology}
\streetaddress{Wuhan, China}
  \city{Wuhan}
  \country{China}
}
\email{haopengfang@hust.edu.cn}

\author{He Tang}
\authornote{Corresponding author: He Tang (E-mail: hetang@hust.edu.cn)}
 \affiliation{School of Software Engineering, 
   \institution{Huazhong University of Science and Technology}
  \city{Wuhan}
  \country{China}
}
\email{hetang@hust.edu.cn}

\begin{abstract}
In recent years, saliency ranking has emerged as a challenging task focusing on assessing the degree of saliency at instance-level. Being subjective, even humans struggle to identify the precise order of all salient instances. Previous approaches undertake the saliency ranking by directly sorting the rank scores of salient instances, which have not explicitly resolved the inherent ambiguities. To overcome this limitation, we propose the ranking by partition paradigm, which segments unordered salient instances into partitions and then ranks them based on the correlations among these partitions. The ranking by partition paradigm alleviates ranking ambiguities in a general sense,  as it consistently improves the performance of other saliency ranking models. Additionally, we introduce the Dense Pyramid Transformer (DPT) to enable global cross-scale interactions, which significantly enhances feature interactions with reduced computational burden. Extensive experiments demonstrate that our approach outperforms all existing methods. The code for our method is available at \url{https://github.com/ssecv/PSR}.

\end{abstract}



\settopmatter{printacmref=false} 
\renewcommand\footnotetextcopyrightpermission[1]{}
\maketitle

\begin{figure}
	\begin{center}
		\includegraphics[width=1\linewidth]{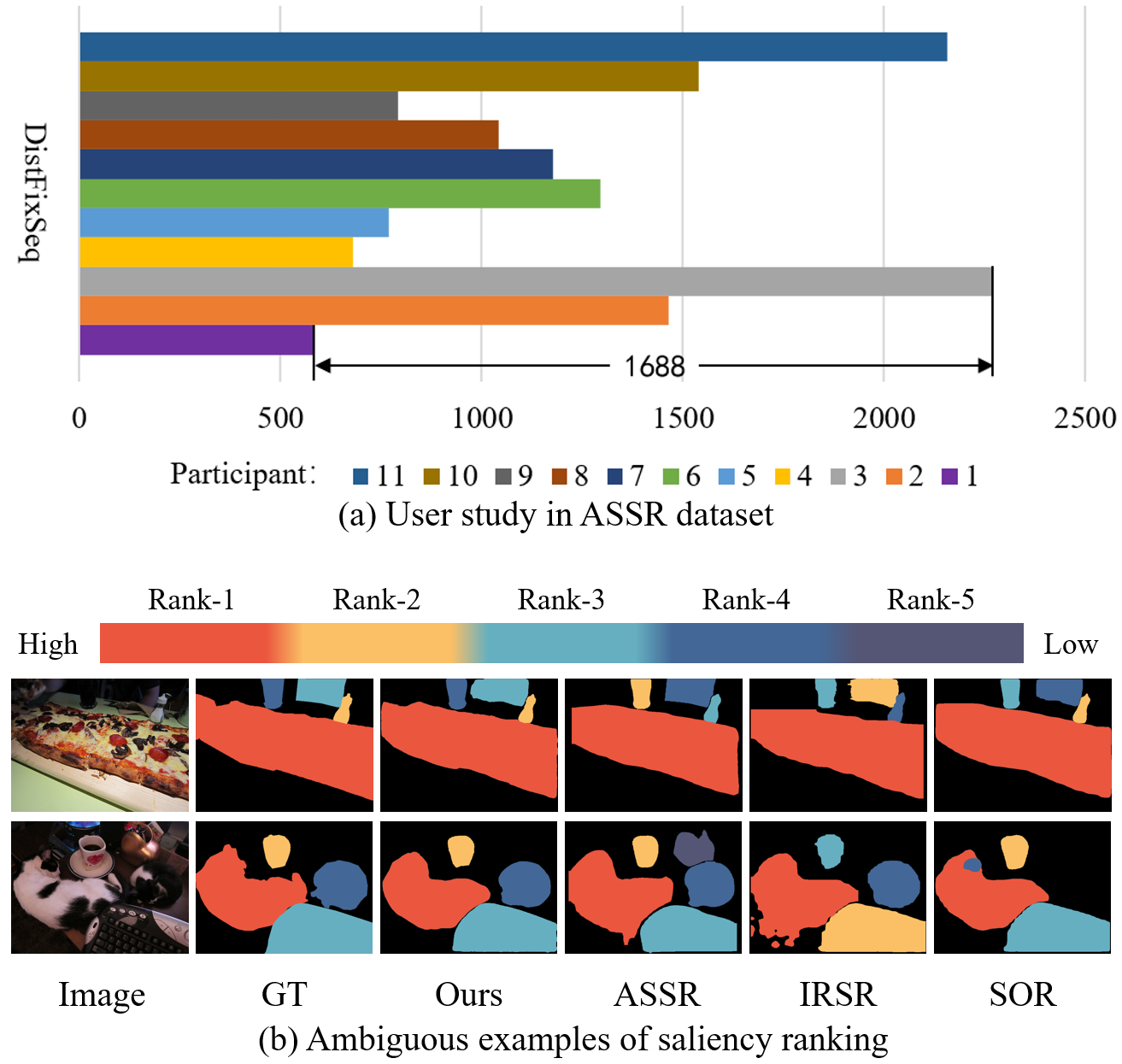}
		\caption{Challenges in saliency ranking. (a) The user study conducted on the ASSR dataset revealed a discrepancy among participants regarding the methodology for determining the degree of saliency. (b) Previous works have not been successful in predicting saliency ranks and masks in ambiguous and complex situations.
		}\label{fig:introduction}
	\end{center}
\end{figure}

\section{Introduction}
Saliency detection is a crucial research area in computer vision. Prior studies have focused on pixel-level salient object detection \cite{liu2010learning,cheng2014global,yang2013saliency,feng2019attentive,zhao2019egnet,wei2020label,zhang2021auto,siris2021scene,zhao2021complementary,gao2020highly} and salient instance segmentation \cite{li2017instance,wu2021regularized,liu2021scg,tian2020weakly,pei2022transformer}. Though these works have achieved promising results, they do not take into account the relative saliency ranks among objects, which is more aligned with the human visual system \cite{seth2022theories}. In order to tackle this problem, Islam \etal{} \cite{islam2018revisiting} introduce a new task called saliency ranking (SR), which not only segments the salient instances from the image but also predicts the relative saliency ranks of these instances. Subsequent works \cite{siris2020inferring,fang2021salient,liu2021instance,tian2022bi} advance the pixel-level prediction \cite{islam2018revisiting} to instance-level prediction. This is promising to be applied in other vision tasks, like person re-identification \cite{chen2022saliency,richardwebster2022doppelganger}, human gaze communication \cite{fan2019understanding} and video conversion \cite{zhu2021horizontal}.

For generating the most consistent ground truth (GT) of saliency ranking, Siris \etal{} \cite{siris2020inferring} compare nine methods via a user study with 11 participants across 2,500 images. Among the methods, DistFixSeq gains the highest number of image picks from participants and being adopted to generate the ground truth of saliency ranking. However, as shown in Fig. \ref{fig:introduction}(a), there is a discrepancy in the number of image picks between participant-1 and participant-3, which reaches 1688 images with a proportion of 67.5\%. This shows that GT of saliency ranking can still be ambiguous even when using the most consistent generating method. This ambiguity may be due to the fact that saliency ranking is difficult even for humans to determine the exact ranks among salient objects without dispute \cite{liu2021instance}.

The intrinsic ambiguity of saliency ranking makes it difficult for modern methods to accurately predict the ranks of salient instances. As shown in the top row of Fig. \ref{fig:introduction}(b), while methods such as ASSR \cite{siris2020inferring}, IRSR \cite{liu2021instance}, and SOR \cite{fang2021salient} can predict reasonable masks of salient instances, their predicted ranks differ among each other, especially for the ones with lower ranks.

\begin{figure}
	\begin{center}
		\includegraphics[width=1\linewidth]{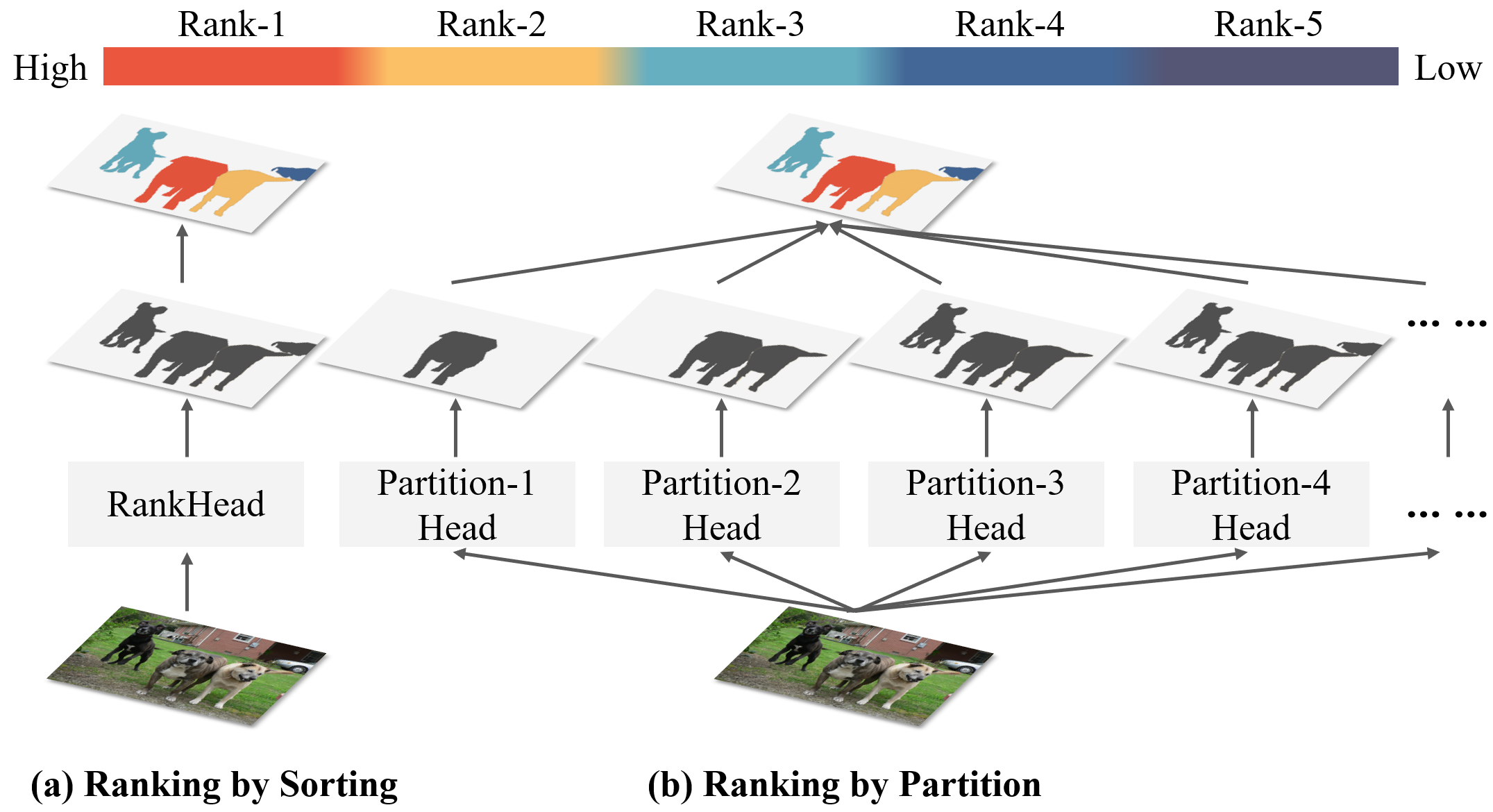}
		\caption{\textbf{(a) Ranking by sorting and (b) ranking by partition.} Ranking by sorting is directly predicting rank of salient instance by sorting the rank scores. Ranking by partition involves identifying saliency partitions of unordered salient instances and ranking the salient instances by the correlations among these partitions.
		}\label{fig:Paradigm}
	\end{center}
\end{figure}

Considering the paradigm of previous works, \cite{siris2020inferring, liu2021instance, fang2021salient, tian2022bi} directly predict ranks of salient instances by sorting the rank scores, \textit{i.e.}, ranking by sorting refer to Fig. \ref{fig:Paradigm}(a). But this paradigm is not designed to explicitly address the inherent ambiguities in saliency ranking, resulting in incorrect assessment for salient instances with inferior ranks, as shown in top row of Fig. \ref{fig:introduction}(b). 

On the other hand, ASSR \cite{siris2020inferring}, IRSR \cite{liu2021instance} and SOR \cite{fang2021salient} are based on sparse interaction of proposal features \cite{he2017mask}. They obtain object proposals at the first stage, then interact these proposals to build relative relationship at the second stage. However, these models select the salient object proposals and discard the background and the objects with lower degrees of saliency. The missing features that not participate in interactions are also useful for saliency ranking. As shown in the bottom row of Fig. \ref{fig:introduction}(b), results of the sparse interaction-based methods ASSR \cite{siris2020inferring}, IRSR \cite{liu2021instance} and SOR \cite{fang2021salient} lead to false positives and false negatives.

In this paper, we tackle the above limitations by alleviating ambiguity and enhancing feature interaction when ranking the salient instances. It is assumed that humans are easier to identify the \textit{entire} $N$ most attractive objects than to determine their exact relative order \textit{one-by-one}. Based on this assumption, we propose a new paradigm to alleviate the ambiguity problem in saliency ranking, namely, ranking by partition. As shown in Fig. \ref{fig:Paradigm}(b), we produce $N$ saliency partitions, where $N$ is equal to the maximum rank of a dataset.

Each partition consists of a set of unordered salient instances that are prioritized as equal to or higher than the corresponding rank. Specifically, the partition-$n$ produces at most $n$ unordered salient instances.
 
To implement this idea, we design partition heads to predict probabilities of $N$ partitions for each instance, and propose Partition to Rank (P2R) to infer the exact rank of each salient instance via the correlations among these saliency partitions. This ranking by partition paradigm is helpful for alleviating saliency ambiguity.

Furthermore, we sufficiently enhance the feature interaction in a global and cross-scale manner. In order to avoid the computational complexity of quadratic to the product of the number of scales and spatial resolution, we design the Dense Pyramid Transformer (DPT), dividing the interaction process into three routes: row attention, column attention and cross-scale attention. 
The proposed DPT outperforms Twin Transformer \cite{guo2021sotr} and Deformable Transformer \cite{zhu2020deformable} with aligned speed.

To sum up, the contributions of this work are as follows: 
\begin{itemize}
    \item To alleviate the ambiguity in saliency ranking, we propose a new paradigm, ranking by partition other than ranking by sorting.
    \item We propose dense pyramid transformer (DPT) that achieves global cross-scale interaction with reduced  computational burden.
    \item We conduct extensive experiments to analyze our approach and verify its superior performance over the state-of-the-art methods.
\end{itemize}

\begin{figure*}
	\begin{center}
		\includegraphics[width=1\textwidth]{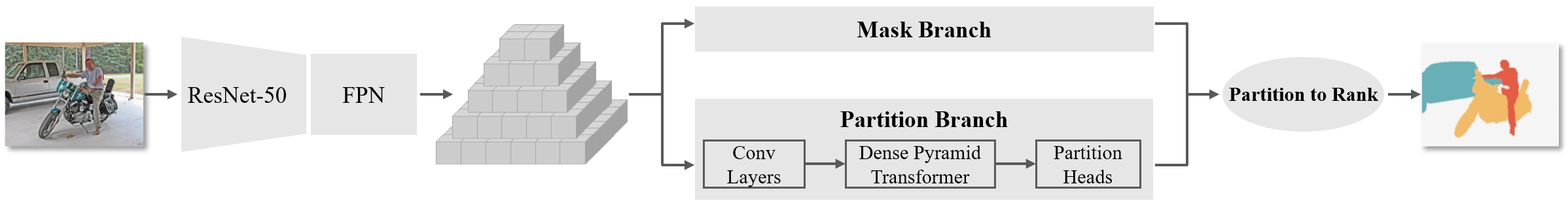}
		\caption{The overall architecture of  the proposed partitioned saliency ranking. 
}\label{fig:architecture}\end{center}
\end{figure*}
\section{RELATED WORK}

\subsection{Saliency Ranking}
Saliency Ranking task is motivated by real-world scenarios where human prioritize certain objects over others. Islam \etal{} \cite{islam2018revisiting} first introduced saliency ranking in the computer vision community. They treated it as a pixel-wise regression problem. Siris \etal{} \cite{siris2020inferring} introduced a new dataset ASSR according to human attention shift, and proposed a method that used both bottom-up and top-down attention mechanisms to predict the ranks and masks of salient instances. This was the first instance-level saliency ranking method, the subsequent works followed this objective. Liu \etal{} \cite{liu2021instance} built another dataset IRSR that depended on the duration of gaze instead of the sequential order. In addition, they designed a graph convolution based network to predict relative saliency ranking and proposed a ranking loss that incorporated rank order of GT. Fang \etal{} \cite{fang2021salient} proposed the first transformer-based position-preserved attention module and used an end-to-end multi-task model that simultaneously performed instance segmentation and saliency ranking. Tian \etal{} \cite{tian2022bi} proposed a bi-directional proposal interaction method with a selective object saliency module. In the domain of video, \cite{wang2019ranking} proposed a technique to rank the saliency of objects based on the relative fixations of human observers. 

By modeling the interaction among proposals and context, the above methods predict mask and relative ranks of salient instances. However, they follow the ranking by sorting paradigm that not adequately address the label ambiguity challenge. In contrast, in this paper, we propose a new ranking by partition paradigm with dense pyramid transformers to alleviate saliency ambiguity and promote cross-scale global feature interaction.

\subsection{Regular Instance Segmentation}
Regular Instance Segmentation (RIS) is a fundamental vision task that predicts all masks and semantic classes of objects in a scene. Mask R-CNN \cite{he2017mask} was a typical instance segmentation method which improved on \cite{ren2015faster} by using FCN \cite{long2015fully} to predict class and mask for each proposal generated by ROIAlign layer. Ma \etal{} \cite{ma2021implicit} introduced a implicit feature refinement module for high-quality instance segmentation. SOLO series \cite{wang2020solo,wang2020solov2} segmented objects by location without box detection. MaskFormer series \cite{cheng2021per,cheng2022masked} predicted a set of binary masks with a single global class label prediction, implementing semantic and instance segmentation in a unified framework. Recently, bipartite matching \cite{carion2020end} based methods \cite{fang2021instances,cheng2022sparse} achieved promising results without NMS in postprocessing. Besides, video instance segmentation also achieved significant improvements with tracklet query,  tracklet proposal \cite{qin2021learning} and generative model \cite{qin2021learning}. RIS and SR are similar on the segmentation task, meanwhile, the difference between them is that the predicted semantic class of RIS is independent among each object rather than the relative rank of saliency in SR. Nevertheless, modern RIS models like Mask R-CNN \cite{he2017mask} and QueryInst \cite{fang2021instances} are adopted by \cite{siris2020inferring,liu2021instance,tian2022bi} as base models for saliency ranking. As the simplification, we select SOLOv2 \cite{wang2020solov2} as our base model. 

\subsection{Salient Instance Segmentation}
Salient Instance Segmentation (SIS) aims to segment individual salient objects rather than pixel-level SOD. Li \etal{} \cite{li2017instance} first introduced the concept of salient instance segmentation and proposed a corresponding dataset containing 1000 images with instance and pixel-level as well as contours annotations. They also proposed a multi-scale network MSRNet to segment salient instances with the help of salient object contours. Subsequent works \cite{fan2019s4net,wu2021regularized,liu2021scg} segmented salient instance in a single-stage manner without using contours annotations. Weakly supervised SIS methods \cite{tian2020weakly,pei2022salient} predicted mask of each salient object without using the instance-level annotations. Recently, Pei \etal{} proposed a method \cite{pei2022transformer} which leveraged the long-range dependencies of transformers and built a new dataset SIS10K.

Considering the maximum number of salient instances, SIS can be regarded as a special case of SR. Since SIS also segment masks of each salient instance as SR, but do not distinguish the saliency degree and ranking order. If the binary (salient/non-salient) classification head of a SIS model is extended to the maximum number of saliency ranking, it can also be served as a satisfactory saliency ranking model.

\section{Problem Formulation}
\subsection{Ranking by sorting} 
Previous saliency ranking methods \cite{siris2020inferring, liu2021instance, fang2021salient, tian2022bi} segment the image $I\in\mathbb{R}^{3\times H\times W}$ into a set of masks, then sort the rank scores of salient instances for ranking, \ie{}, ranking by sorting. This process is depicted in Fig. \ref{fig:Paradigm}(a). We assume that $N$ is equal to the maximum rank of the dataset and $f(\cdot)$ is the forward process of the network. Let $M = \{m_{i}|m_{i}\in\left\{0,1\right\}^{H\times W}\}_{i=1}^N$, denote the masks of instances, and $R = \{r_{i}|r_{i} \in \{1, 2, ..., N\}\}_{i=1}^N$, denote the ranks of instances. The prediction including mask and saliency rank of each instance is formulated as
 \begin{equation}
 \begin{split}
<m_i, r_\textit{i}> = f(I).
 \end{split}
 \end{equation}
However, as shown in Fig. \ref{fig:introduction}(b), ranking by sorting based methods \cite{siris2020inferring, liu2021instance, fang2021salient} fail to predict masks and ranks of salient instances with lower ranks.

\subsection{Ranking by partition} 
Generally speaking, it is less complicated for individuals to select $N$ most salient instances concurrently than to identify the precise sequential ordering of salient instances. Inspired by this perspective, we propose a novel paradigm, \ie{}, ranking by partition which reduces the ambiguity of ranking each salient instance. 

As delineated in Fig. \ref{fig:Paradigm}(b), each partition consists of a set of unordered salient instances that are prioritized as equal to or higher than the corresponding rank. Let $P = \{p_{n}\}_{n=1}^{N}$ denotes the probabilities of each instance belonging to the $N$ partitions. The outputs of the network comprise predictions in the form of mask and partition probabilities for each instance, which can be formulated as
 \begin{equation}
 \begin{split}
<M, p_1, p_2, ...,p_N > = f(I).
 \end{split}
 \end{equation}
We propose the Partition to Rank (P2R) to infer the rank of each instance by parsing the correlations among these partitions. The parsing result is formulated as
 \begin{equation}
 \begin{split}
<m_i, r_\textit{i}=n> = P2R(M, p_1, p_2,..., p_n).
 \end{split}
 \end{equation}

\section{Methodology}
\subsection{Overall architecture}
The overall architecture is depicted in Fig. \ref{fig:architecture}. The network first utilizes a convolutional backbone, ResNet-50 \cite{he2016deep}, along with a feature pyramid network (FPN) \cite{lin2017feature}. Given an input image $I\in\mathbb{R}^{3\times H\times W}$, we extract multi-scale features $C = \{c_{i}\}_{i=2}^6$ from the FPN.

The multi-scale features $C$ are partitioned into grids, with the size of the grids varying across scales. The grids can be formulated as $G = \{g_{i}|g_{i}\in\mathbb{R}^{E\times s_{i}\times s_{i}}\}_{i=2}^6$, where $E$ denotes the number of channels and $\{s_{i}\}_{i=2}^6$ signifies the side length of grids. The grids $G$ serves as the input to the mask branch and partition branch.

Following SOLOv2 \cite{wang2020solov2}, we adopt dynamic instance segmentation as our mask branch. Leveraging the grids $G$, dynamic convolution kernels are computed. We then integrate the features from FPN to obtain a new global feature map. The instance masks are obtained by performing dynamic convolutions on the global feature map.

\begin{figure}
	\begin{center}
		\includegraphics[width= \linewidth]{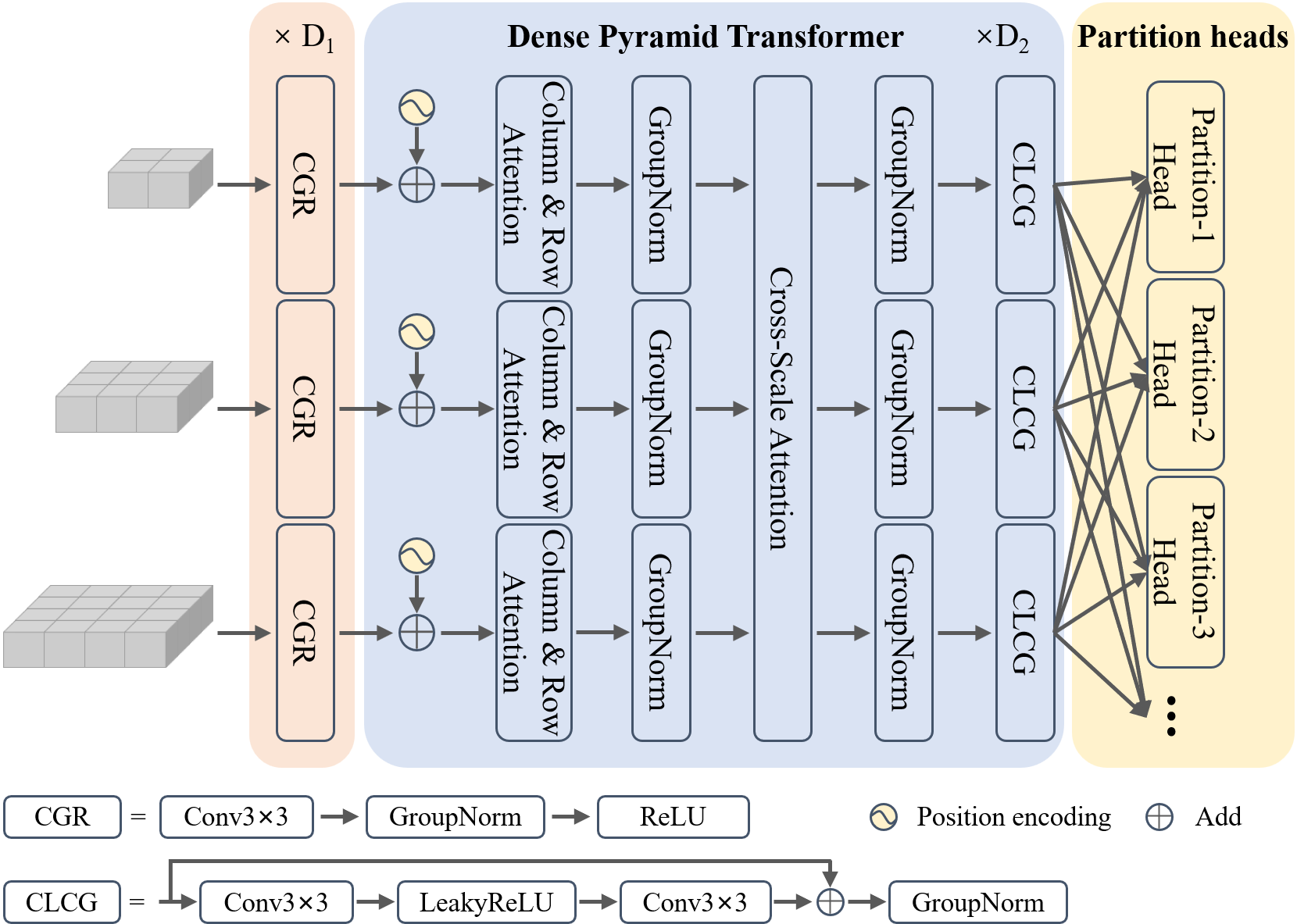}
		\caption{\textbf{Partition branch.} It includes convolutional layers, dense pyramid transformer layers and partition heads. 
  For graphical perspicuity, the illustration omits two scales of feature maps.
		}\label{fig:Rank Branch}
	\end{center}
\end{figure}

To predict saliency partitions, we design a partition branch consisting of three components as delineated in Fig. \ref{fig:Rank Branch}. The first component comprises convolutional layers which perform preliminary processing on $G$. The second part, DPT, enables comprehensive interaction among all grid cells across scales. Finally, there are $N$ partition heads to predict partition probabilities for each instance utilizing the outputs of the DPT.

\subsection{Partition branch} 
\subsubsection{Convolutional Layers.}
Previous works \cite{liu2021instance,fang2021salient,tian2022bi} provide the evidence that feature interaction is crucial for saliency ranking. 
We attempt to establish global cross-scale feature interaction with transformers.

According to \cite{park2022vision}, self-attention and convolutions are complementary. Hence, we adopt convolutional layers before global cross-scale feature interaction for harmonization. The input to the convolutional layers is the grids $G$. The process of harmonizing is described as CGR in Fig. \ref{fig:Rank Branch}. The output of CGR is formulated as
 \begin{equation}
 \begin{split}
\hat{G} = ReLU(GN(Conv(G))),
 \end{split}
 \end{equation}
where $\hat{G}$ denote the harmonized feature.

\begin{figure}
	\begin{center}
		\includegraphics[width= \linewidth]{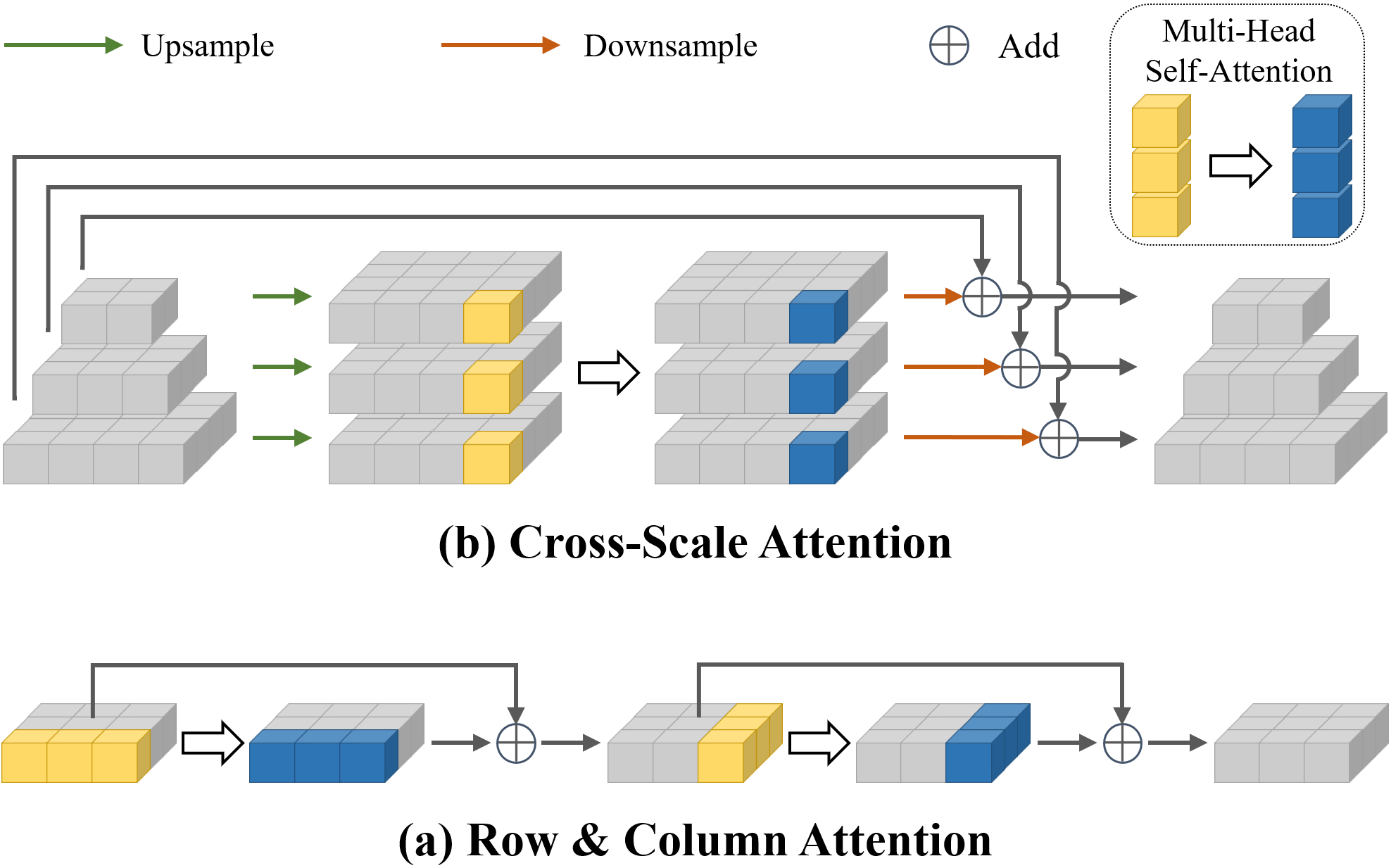}
		\caption{Dense pyramid transformer consists of (a) Row \& Column Attention and (b) Cross-Scale Attention. 
  For graphical perspicuity, the illustration omits some scales of feature maps.
		}\label{fig:cross-scale attention}
	\end{center}
\end{figure}

\subsubsection{Dense Pyramid Transformer.}
To achieve global cross-scale feature interaction, one feasible approach is to flatten all features across scales and concatenate them before enabling interaction through multi-head self-attention (MHSA), refer to all-scale transformer in this paper. However, if transformer layers are designed this way, the computational complexity per layer would be $O((\mathcal{S}\times H\times W)^2)$, where $\mathcal{S}$ is the number of scales.

We find there are two elegant transformers which achieve better interaction with less computation. One is twin transformer \cite{guo2021sotr} which decomposes 2D attention into column and row attentions. Another is deformable transformer \cite{zhu2020deformable} which can interact with cross-scale features efficiently. Inspired by them, we propose a novel transformer architecture, dense partition transformer (DPT), which divides the interaction process into three components: row attention, column attention and cross-scale attention. The proposed DPT first facilitates comprehensive interaction among features within the same scale through row and column attentions. It then establishes cross-scale attention to promote interaction between features at different scales. This reduces the computational complexity to $O(\mathcal{S}\times H\times W^2 + \mathcal{S}\times H^2\times W+\mathcal{S}^2\times H\times W) $, while preserving the comprehensiveness of interaction.

The detailed structure of DPT is shown in Fig. \ref{fig:Rank Branch}. We first add position encoding to $\hat{G}$ and acquire positioned feature $F$. Subsequently, we utilize the row and column attention, as shown in Fig. \ref{fig:cross-scale attention}(a), to enable interaction within each scale of $F$. In each scale, MHSA is first applied within each row of features, which are then combined with the original features through residual connections. This procedure is then repeated within each column of features. In summary, the process of row and column attentions is described as
  \begin{equation}
 \begin{split}
F^R_{:,y,z} = MHSA(F_{:,y,z}) + F_{:,y,z}, 
 \end{split}
 \end{equation}
  \begin{equation}
 \begin{split}
F^{RC}_{x,:,z} = MHSA(F^R_{x,:,z}) + F^R_{x,:,z},
 \end{split}
 \end{equation} where $F^R$ and $F^{RC}$ are the outputs of row attention and row \& column attention respectively. We normalize $F^{RC}$ into $F^{'}$ by group normalization \cite{wu2018group}. In cross-scale attention, we need a pre-processing step to guarantee that the gird sizes of each level are uniform, which enables cross-scale feature interaction in the subsequent steps. As can be seen in Fig. \ref{fig:cross-scale attention}(b), we upsample the girds to the size of the largest one. After applying MHSA to features across different scales at equivalent locations, we restore the grid of each scale to their original shape and add a residual connection. The output of cross-scale attention can be expressed as
  \begin{equation}
 \begin{split}
F^{RCC}_{x,y,:} = Downsample(MHSA(Upsample(F^{'}_{x,y,:}))) + F^{'}_{x,y,:},
 \end{split}
 \end{equation} where $F^{RCC}$ is the output of cross-scale attention. Then we normalize $F^{RCC}$ into $F^{''}$ by group normalization. The terminal component of DPT adopts a convolutional neural network architecture denoted as CLCG. CLCG comprises two 3x3 convolutional layers connected by a LeakyReLU and followed by group normalization. Prior to the group normalization layer, residual connection is implemented. The output of DPT $\hat{F}$ can be computed as
  \begin{equation}
 \begin{split}
\hat{F} = GN(Conv(LeakyReLU(Conv(F^{''}))) + F^{''}).
 \end{split}
 \end{equation}
The addition of a convolution operation has been proposed as a useful complement to the attention mechanism, as it can better capture local information and enhance feature representation.

\begin{figure}
	\begin{center}
		\includegraphics[width= \linewidth]{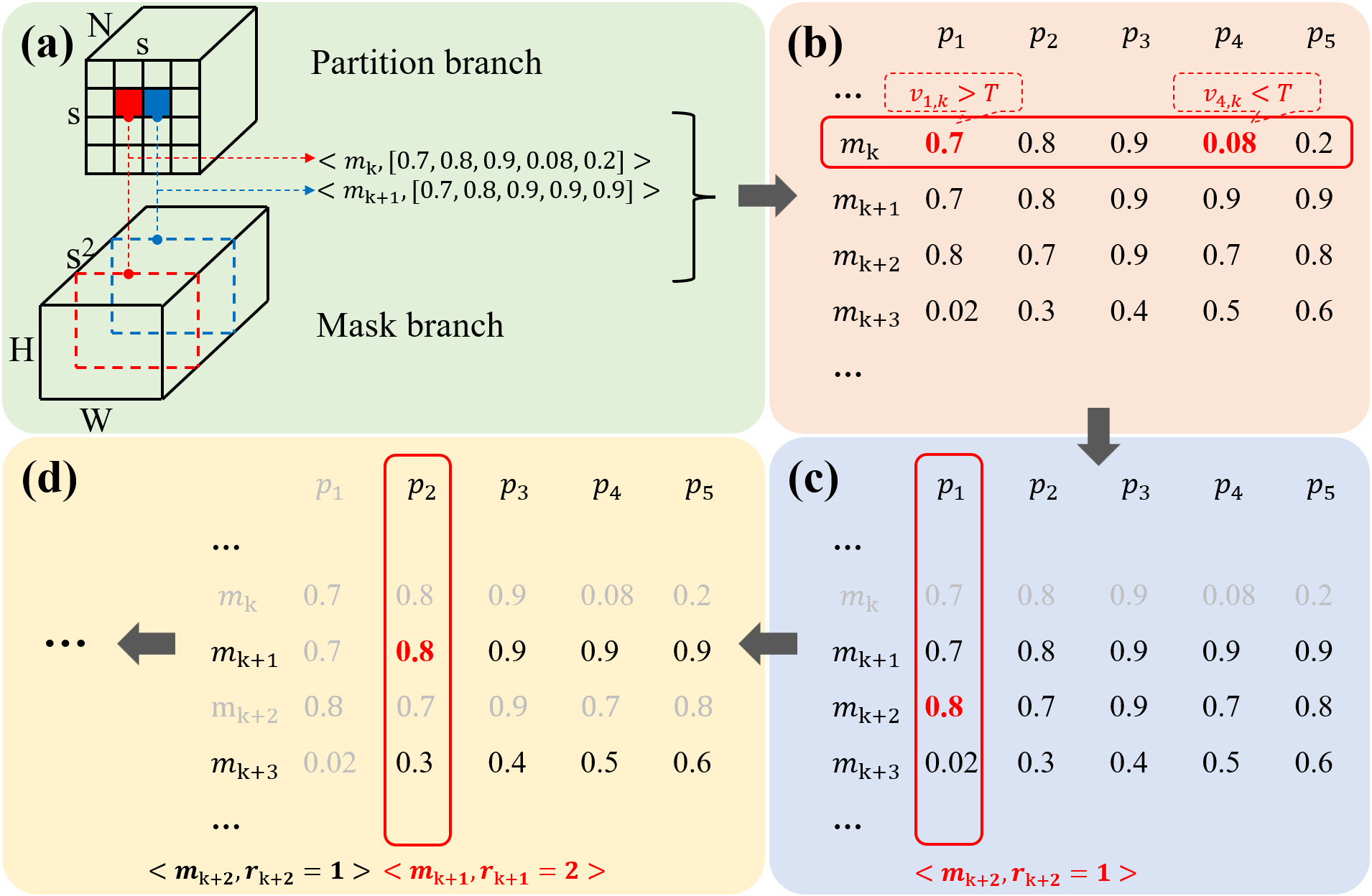}
		\caption{The illustration of our partition to rank (P2R), the ranks are ranging from 1 to 5. For graphical perspicuity, the illustration omits some scales of feature maps. (a) The associated output of partition branch and mask branch; (b)  ambiguity alleviation of saliency ranking; (c) rank-1 instance selection; (d) instance of other ranks selection.
		}\label{fig:P2R}
	\end{center}
\end{figure}

\subsubsection{Partition Heads.} 
As can be seen in Fig. \ref{fig:Paradigm}, the partition heads of our proposed approach are designed as $N$ binary saliency classification heads, where $N$ represents the maximum number of ranks. 
Each partition head is a convolutional layer of $E\times 3 \times 3 \times 1$, where $E$ denotes the number of input channels, it connects to all scales of gird cells from DPT. The output of partition heads is $P=\{p_i|p_i=[v_{i,1},...,v_{i,{\sum_{j=2}^{6}}s_j\times s_j}]^T\}_{i=1}^N$ , if we distinguish the salient instances into $N$ ranks, where each logit $v\in[0,1]$ and $s_j$ is the side length of each grid.
In this way, each partition predicts a set of unordered salient instances that ranks equal or higher than the partition index.

\begin{table*}[t]
\centering{
\caption{Performance comparison with 16 state-of-the-art instance-level segmentation methods on two datasets. Smaller $MAE$, larger $SA-SOR$ and $SOR$ indicates better performance. The best result is in \textbf{bold} and the second is \underline{underlined}.}
\label{table: SR benchmark}
\setlength{\tabcolsep}{3.5mm}{
    \begin{tabular}{c|c|c|c|c|c|c|c|c|c}
    \toprule[1pt]
    \multirow{2}{*}{Method} & \multirow{2}{*}{Task} & \multicolumn{3}{c|}{ASSR Dataset} & \multicolumn{3}{c}{IRSR Dataset} & \multicolumn{1}{|c|}{\multirow{2}{*}{\#Para.(M)}} & \multirow{2}{*}{FPS} \\
    \cline{3-8}
                    &               & MAE$\downarrow$ & SA-SOR$\uparrow$ & SOR$\uparrow$    & MAE$\downarrow$  & SA-SOR$\uparrow$  & SOR$\uparrow$    &         &         \\
    \hline
    \hline
    Mask2Former \cite{cheng2022masked}   & \multirow{8}{*}{RIS}    & 0.085 & 0.694  & 0.862 & 0.097                & 0.493                & 0.833 & 44.0  & 30.1 \\
    SparseInst \cite{cheng2022sparse}    &                         & 0.103 & 0.669  & 0.858 & 0.112                & 0.464                & 0.848 & 31.6  & 50.4 \\
    SOTR \cite{guo2021sotr}         &                         & 0.092 & 0.682  & 0.867 & 0.111                & 0.470                & 0.823 & 63.1  & 11.3 \\
    SOLO \cite{wang2020solo}          &                         & 0.117 & 0.643  & 0.827 & 0.125                & 0.465                & 0.820 & 36.1  & 14.7 \\
    SOLOv2 \cite{wang2020solov2}        &                         & 0.096 & 0.676  & 0.848 & 0.114                & 0.479                & 0.831 & 45.0  & 22.0  \\
    Mask R-CNN \cite{he2017mask}    &                         & 0.128 & 0.624  & 0.731 & 0.131                & 0.437                & 0.813 & 43.8  & 22.8 \\
    Cascade R-CNN \cite{cai2018cascade} &                         & 0.109 & 0.661  & 0.809 & 0.121                & 0.452                & 0.825 & 76.8  & 16.2 \\
    QueryInst \cite{fang2021instances}     &                         & 0.088 & 0.698  & 0.832 & 0.105                & 0.519                & 0.833 & 172.2 & 15.7 \\
    \midrule
    RDPNet \cite{wu2021regularized}        & \multirow{4}{*}{S/C IS} & 0.119 & 0.641  & 0.826 & 0.134                &   0.491              & 0.809  & 44.3  & 20.7   \\
    SCG \cite{liu2021scg}           &                         & 0.107 & 0.659  & 0.842 & 0.123                &   0.518              & 0.816 & 47.9  & 7.2  \\
    OQTR \cite{pei2022transformer}          &                         & 0.094 & 0.677  & 0.865 & 0.116                &   0.535              & 0.831 & 43.1 & 25.4  \\
    OSFormer \cite{pei2022osformer}      &                         & 0.118 & 0.626  & 0.831 & 0.112                & 0.438                & 0.817 & 46.6 & 26.3 \\
    \midrule
    ASSR \cite{siris2020inferring}          & \multirow{5}{*}{SR}     & 0.104 & 0.661  & 0.787 & 0.134                & 0.377                & 0.710 & 43.9  & 22.1  \\
    IRSR \cite{liu2021instance}          &                         & 0.105 & 0.705  & 0.813 & \underline{0.088}                & \underline{0.564}                & \underline{0.806} & 102.4 & 18.5 \\
    SOR \cite{fang2021salient}           &                         & \underline{0.083} & 0.717  & 0.836 & 0.091                & 0.543                & 0.797 & 100.0  &  36.1  \\
    OCOR \cite{tian2022bi}    &                         & 0.085 & \underline{0.723}  & \underline{0.877} & -                    & -                    & -     & -     & -     \\
    PSR (proposed)           &                         & \textbf{0.075} & \textbf{0.738}  & \textbf{0.892} & \textbf{0.080}                & \textbf{0.651}                & \textbf{0.878} & 50.9  & 20.9  \\
    \midrule
    \midrule
    OCOR-Swin-L \cite{tian2022bi}   & \multirow{2}{*}{SR}     & \underline{0.078} & \underline{0.738}  & \underline{0.904} & \underline{0.079}                & \underline{0.578}                & \underline{0.834} & -     & -     \\
    PSR-Swin-L    &                         & \textbf{0.071} & \textbf{0.746}  & \textbf{0.915} & \textbf{0.075}                &   \textbf{0.664}              & \textbf{0.890}  & 225.3  & 7.7  \\
    \bottomrule[1pt]
    \end{tabular}
}
}
\end{table*}

\subsection{Partition to Rank}
Fig. \ref{fig:P2R} delineates the process by which the outputs from the mask branch and partition branch are associated to engender the final prediction of saliency ranking. 
\begin{enumerate}[(a)]
  \item Association. We associate the predicted instance masks and their corresponding partition probabilities on their reference grid cells. This association yields tuples that compose each row of the partition matrix, as illustrated in Fig. \ref{fig:P2R}(a).
  \item Ambiguity alleviation. Presented in Fig. \ref{fig:P2R}(b), we discard one instance when 1) the probability of partition-$i$ is below the threshold $T$ as $v_{i,k}<T$; and 2) there exists a probability of partition-$j$ exceed $T$ as $v_{j, k} \geq T$; and 3)  $j<i$. The threshold $T$ is set to 0.3 in this paper for best performance. 
  \item Rank-1 selection. We consider the instance with the largest probability of partition-1 as rank-1, \textit{e.g.}, $m_{k+2}$ with a 0.8 probability of partition-1, as shown in Fig. \ref{fig:P2R}(c). Non-Maximum Suppression (NMS) is then performed on remaining instances, suppressing the instances with an Intersection over Union (IoU) exceeding 0.5.
  \item Other ranks selection. In Fig. \ref{fig:P2R}(d), we discard the rank-1 instance from the partition matrix and use the same approach as rank-1 selection to process the remaining instances. This process is repeated until all $N$ salient instances are selected or all instances in the partition matrix are discarded.
\end{enumerate}
 Finally, we apply a post-process for the predictions of saliency ranking. A threshold of 0.5 is adopted to convert the predicted soft masks into binary masks.

\subsection{Loss Function}

In order to obtain the ground truth of the partitions, for each instance mask, we convert the corresponding rank label into a boolean vector of length $N$. The $n$-th boolean value in the vector signifies whether the salient instance is prioritized equal to or higher than rank-$n$. 

The training loss function is defined as follows
 \begin{equation}
 \begin{split}
L =  \sum_{n=1}^{N} \lambda_{partition} L_{partition}  + \lambda_{mask}L_{mask}.
 \end{split}
 \end{equation}
We train the partition branch by $L_{partition}$ which is computed by focal loss \cite{lin2017focal}. $L_{mask}$ is computed by Dice loss for segmentation. In addition, $\lambda_{partition}$ and  $\lambda_{mask}$ are the coefficients for partition loss and mask loss.

\section{EXPERIMENTS}

\subsection{Datasets and Evaluation Metrics}
\subsubsection{Datasets.} 
We conduct experiments on two publicly accessible datasets ASSR \cite{siris2020inferring} and IRSR \cite{liu2021instance}. The ASSR dataset employs mouse tracking to collect gaze data, ranks saliency by the first five unique fixated objects. The dataset has 7,464 training, 1,436 validation, and 2,418 test images. The IRSR dataset discards over or under segmented images and those containing over eight or under two salient instances. This dataset includes 6,059 training and 2,929 test images.

 \begin{figure*}
	\begin{center}
		\includegraphics[width=1\linewidth]{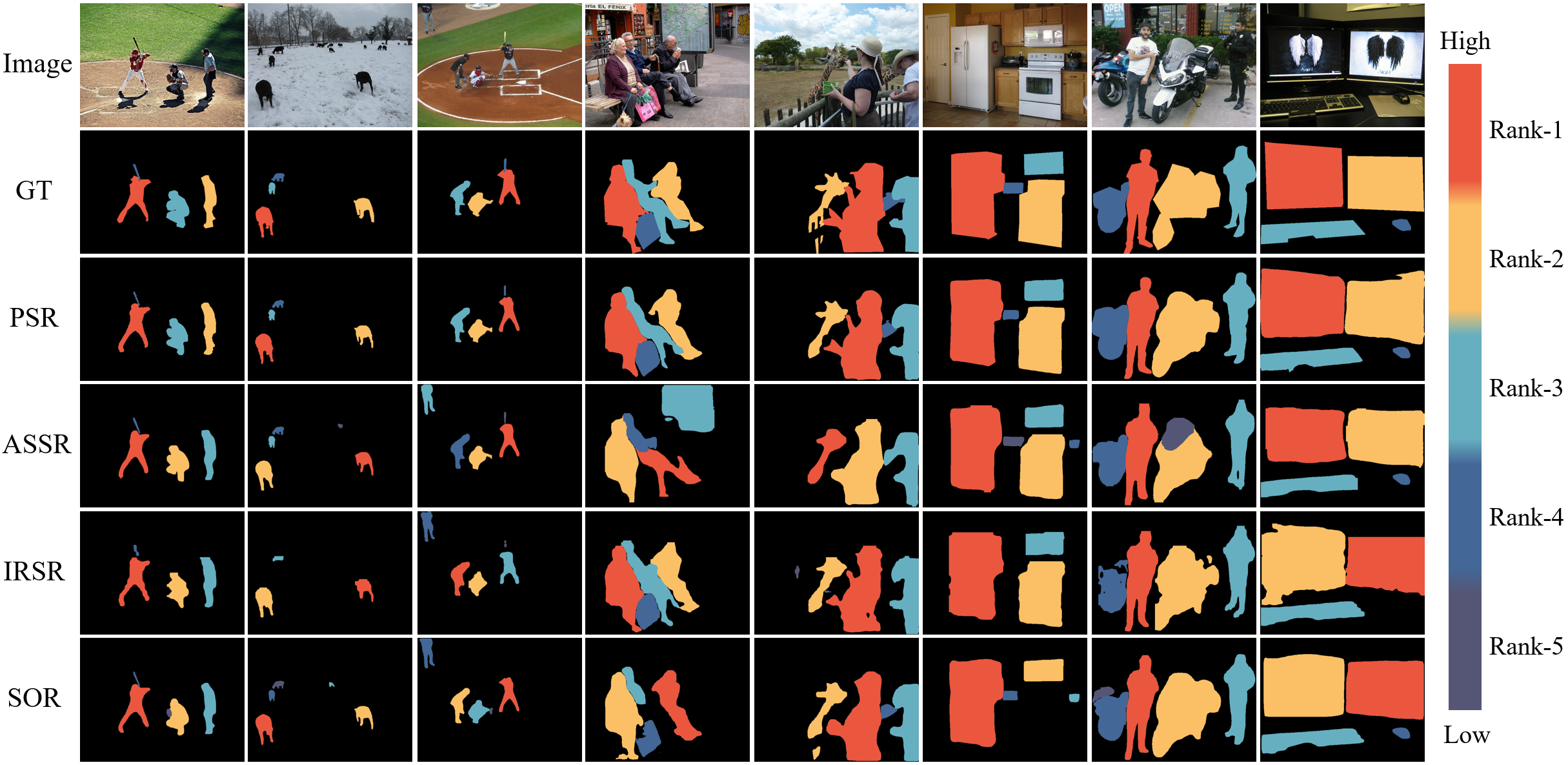}
		\caption{Visual comparison between the proposed PSR and other instance-level saliency ranking methods. Our PSR improves both mask and ranking precision compared to ASSR \cite{siris2020inferring}, IRSR \cite{liu2021instance} and SOR \cite{fang2021salient}.
		}\label{fig:compare}
	\end{center}
\end{figure*}

\subsubsection{Evaluation Metrics.} 
We utilize the identical evaluation methods as \cite{siris2020inferring, liu2021instance} for an equitable comparison, namely Salient Object Ranking (SOR), Segmentation-Aware SOR (SA-SOR), and Mean Absolute Error (MAE). SOR computes the Spearman's rank-order correlation between the predictions and GT. On the other hand, SA-SOR filters the objects at the instance-level using the mask Intersection over Union (IoU) and employs the Pearson correlation coefficient to measure the linear correlation between the prediction and GT. The MAE metric compares the average per-pixel difference between the predictions and GT, accounting for both segmentation and ordering quality as a comprehensive metric. 

\subsection{Implementation Details}
Following preceding works \cite{islam2018revisiting, siris2020inferring, liu2021instance, fang2021salient}, we adopt ResNet-50 \cite{he2016deep} weights pretrained on the MS-COCO \cite{lin2014microsoft} 2017 training split and resize input images to $640\times480$. Besides, our model is trained on the ASSR dataset training set. The partition branch comprises 3 convolutional layers and 3 transformer layers. The weights of the partition loss and mask loss are 1 and 3, respectively. Furthermore, we implement the stochastic gradient descent (SGD) optimizer with a learning rate of 2.5e-5 and employ a warm-up strategy in the initial 1,000 iterations. We train the model for 60 epochs with a batch size of 4 on a NVIDIA RTX 3090 GPU. We perform multi-step decay with a decay factor of 1e-4 at the 42nd and 54th epochs, respectively.

\subsection{Comparisons with the State-of-the-arts}
\subsubsection{Quantitative Comparison.} 
As shown in \tabref{table: SR benchmark}, our model is compared with 16 state-of-the-art (SOTA) methods, their original tasks include regular instance segmentation (RIS) \cite{cheng2022masked, cheng2022sparse, guo2021sotr, wang2020solo, wang2020solov2, he2017mask, cai2018cascade, fang2021instances}, salient/camouflage instance segmentation (S/C IS) \cite{wu2021regularized, liu2021scg, pei2022transformer, pei2022osformer} and saliency ranking (SR) tasks \cite{siris2020inferring, liu2021instance, fang2021salient, tian2022bi}. Unless otherwise specified, all experiments reported in Tab. \ref{table: SR benchmark} are conducted using the ResNet-50 backbone to ensure fair comparison. We use '-' to indicate unavailable data due to the code not being open-sourced.

On the ASSR dataset, the proposed PSR surpasses SOTAs by 0.008, 0.015 and 0.015 in terms of MAE, SA-SOR and SOR metrics, respectively. For the IRSR dataset, PSR surpasses state-of-the-arts by 0.008, 0.087 and 0.072 in terms of MAE, SA-SOR and SOR metrics, respectively. Specially, the last two rows of Tab. \ref{table: SR benchmark} present a comparison between the proposed PSR and the latest OCOR \cite{tian2022bi} using the Swin-L \cite{liu2021swin} backbone. All the results indicate that PSR outperform previous methods in terms of all three metrics on both two datasets. Despite including multiple partition heads and the DPT, our FPS only drops by 1.1 compared to the base model, approaching the speed of ASSR and IRSR.

\subsubsection{Qualitative Comparison.} 
In \figref{fig:compare}, we present the visualization results for qualitative analysis between PSR and other instance-level saliency ranking methods include: ASSR \cite{siris2020inferring}, IRSR \cite{liu2021instance} and SOR \cite{fang2021salient}. Our method produces salient instance masks and ranks closer to GT when dealing with objects with ambiguous rank levels in complex scenes. For example, in the first column, though all models predict correct masks of salient instances, ranks of the right two people are hard to determine, only PSR produces the correct ranking order. In the second column, the size of sheep is variant, previous works do not consider global cross-scale feature interaction and thus gets the incorrect ranking order.

 \begin{figure}
	\begin{center}
		\includegraphics[width=1\linewidth]{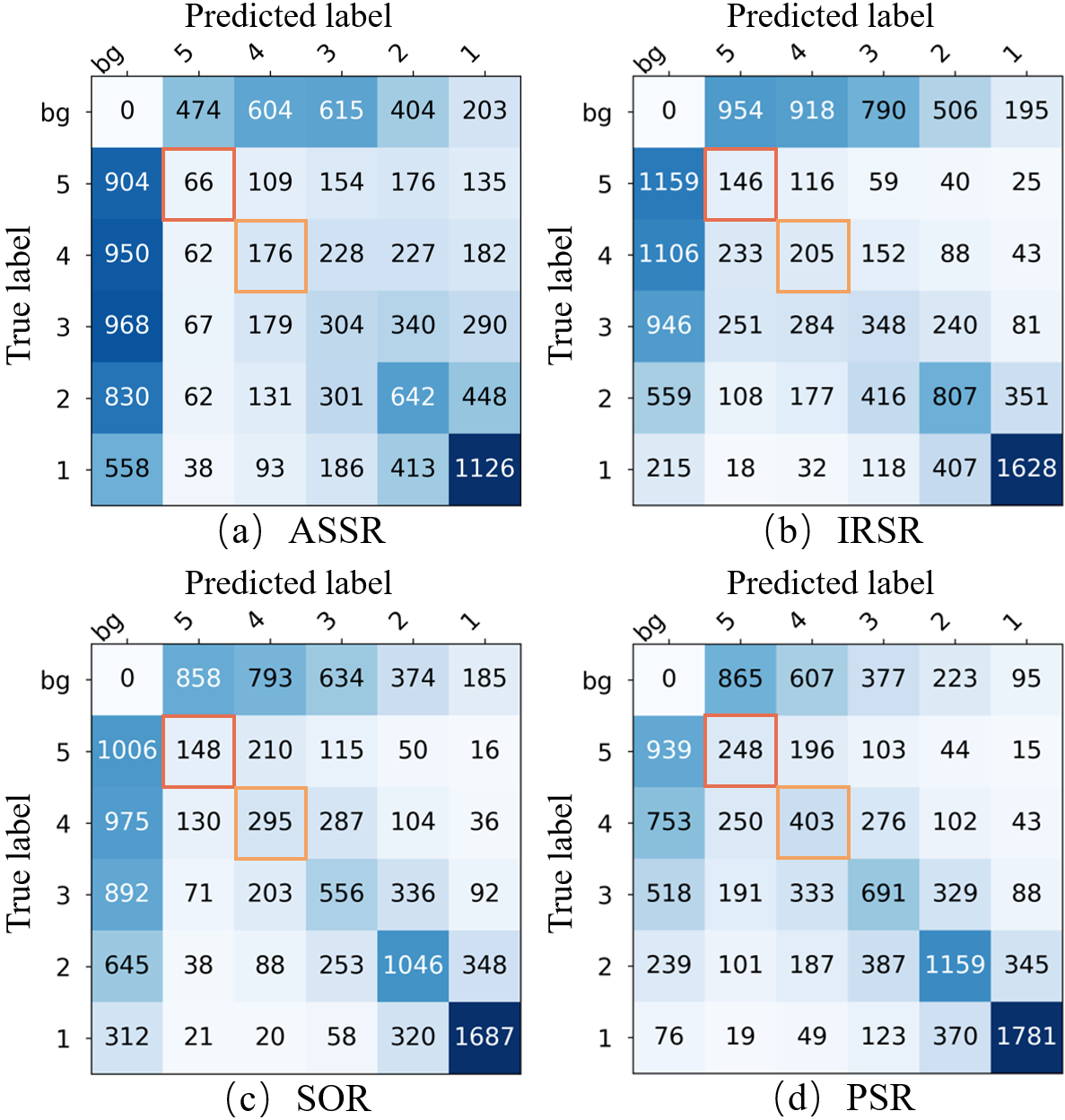}
		\caption{Confusion matrix of the proposed PSR and other instance-level saliency ranking methods on ASSR dataset.
		}\label{fig:confusion}
	\end{center}
\end{figure}
\subsubsection{Confusion Matrix Comparison.}
The confusion matrices of the proposed PSR and other instance-level saliency ranking methods \cite{siris2020inferring,liu2021instance,fang2021salient} are reported on Fig. \ref{fig:confusion}. The principal diagonal line of each matrix reflects the number of true positives for each method. The proposed PSR obtains a summation of 4282, larger than that of 2314, 3134 and 3732 for ASSR \cite{siris2020inferring}, IRSR \cite{liu2021instance} and SOR \cite{fang2021salient} respectively. More importantly, with regard to the rank-5 and rank-4, the number of true positives of PSR reaches 248 and 403 respectively, achieving a relative improvement with 67.6\% and 36.6\% compared to SOR \cite{fang2021salient}.  As the experimental phenomenon, we can see that 1) our PSR outperforms previous models in terms of the overall accuracy; 2) specially, PSR surpass previous methods by a large margin on difficult samples, such as rank-5 and rank-4.

\subsection{Ablation Studies}

\subsubsection{Attention-based Interaction Modules.} The twin transformer \cite{guo2021sotr} lacks of cross-scale interaction and the deformable transformer \cite{zhu2020deformable} lacks global interaction. Hence, we design the DPT to enable comprehensive global cross-scale interaction. We conduct experiments to compare with twin transformer, deformable transformer and all-scale transformer to demonstrate the effectiveness of DPT.

As depicted in the top section of \tabref{table:Ablation of attention}, we attempt to align the FPS by adjusting the number of layers in the twin transformer and deformable transformer. Among the three methods with similar inference speed, the network employing DPT achieved the best performance. 

As shown in the bottom part of \tabref{table:Ablation of attention}, we align the number of transformer layers with 3.
Though it reduces the computational burden, the proposed DPT outperform all-scale transformer in terms of MAE, SA-SOR, SOR, as well as FPS.

\begin{table}[t]
\caption{Comparing the performance of different transformer-based interaction modules.}
\label{table:Ablation of attention}
\setlength{\tabcolsep}{1.6mm}
    \begin{tabular}{c|c|c|c|c|c}
    \toprule[1pt]
    Transformers  & \#Layers & MAE$\downarrow$ & SA-SOR$\uparrow$ & SOR$\uparrow$    & FPS$\uparrow$  \\
    \hline
    Twin \cite{guo2021sotr}          & 4        & 0.080  & 0.719  & 0.883  & 20.7 \\
    Deformable \cite{zhu2020deformable}    & 4        & 0.080  & 0.720  & 0.877  & \textbf{21.2} \\
    Dense Pyramid & 3        & \textbf{0.075}  & \textbf{0.738}  & \textbf{0.892}  & 20.9 \\
    \hline
    \hline
    All-scale     & 3        & 0.079  & 0.726  & 0.885 & 13.1     \\
    Dense Pyramid & 3        & \textbf{0.075}  & \textbf{0.738}  & \textbf{0.892}  & \textbf{20.9} \\
    \bottomrule[1pt]
    \end{tabular}
\end{table}

\subsubsection{Layer Variations in Partition Branch.} We exam the impact of different numbers of convolutional layers and transformer layers in partition branch. As shown in \tabref{table:Ablation of DPI}, the network achieves better performance when convolutional layers are followed by transformer layers. We ultimately adopt a setting of 3 convolutional layers and 3 transformer layers because it achieves best performance.

\begin{table}[t]
\caption{Comparing the performance of layer variations of convolutions and transformers in partition branch.}
\label{table:Ablation of DPI}
\setlength{\tabcolsep}{1.5mm}
    \begin{tabular}{c|c|c|c|c}
    \toprule[1pt]
    \#Conv Layers & \#Transformer Layers & MAE$\downarrow$ & SA-SOR$\uparrow$ & SOR$\uparrow$    \\ \hline
    -             & 6                    & 0.080           & 0.725  & 0.877  \\
    3             & 3                    & \textbf{0.075}  & \textbf{0.738}  & \textbf{0.892}  \\
    6             & -                    & 0.083           & 0.720  & 0.873  \\
    \bottomrule[1pt]
    \end{tabular}
\end{table}

\subsubsection{Contribution of Different Components.} To ascertain the efficacy of the Cross-Scale Attention (C-S Attn) and Row \& Column Attention (R\&C Attn) in DPT, as well as the proposed ranking by partition paradigm (Partition), we devise ablation experiments on the constituent components. The results of the ablation study are presented in \tabref{table:Ablation of component}, with SOLOv2 \cite{wang2020solov2} serving as the base model, listed in the first column. Comparing to the baseline, by leveraging both Cross-Scale and Row \& Column Attentions, the performance improved by 0.010, 0.021 and 0.018 in terms of MAE, SA-SOR and SOR respectively. Employing only ranking by partition also improves the MAE, SA-SOR and SA-SOR by 0.016, 0.037 and 0.023, as this new paradigm mitigates the ambiguity challenges of ordering scenes. The overall results indicate that our method achieves the best performance when all components are included, which demonstrates the effectiveness and necessity of each component.

\begin{table}[t]
    \caption{Comparing the contribution of different components.}
    \label{table:Ablation of component}
    \setlength{\tabcolsep}{2mm}
    \begin{tabular}{c|c|c|c|c|c}
    \toprule[1pt]
    C-S Attn & R\&C Attn  & Partition & MAE$\downarrow$   & SA-SOR$\uparrow$   & SOR$\uparrow$   \\ \hline
            &           &           & 0.096  & 0.676    & 0.848 \\
            & $\surd$   &           & 0.092  & 0.688    & 0.853 \\
    $\surd$ & $\surd$   &           & 0.086  & 0.697    & 0.866 \\
            &           & $\surd$   & 0.080  & 0.713    & 0.871 \\
            & $\surd$   & $\surd$   & 0.077  & 0.724    & 0.884 \\
    $\surd$ & $\surd$   & $\surd$   & \textbf{0.075}  & \textbf{0.738}    & \textbf{0.892} \\
    \bottomrule[1pt]
    \end{tabular}
\end{table}

\subsection{Generalization Experiment of Ranking by Partition}

\begin{table}[t]
    \caption{Generalization Experiment of Ranking by Partition} 
    \label{table:Ablation of partition}
    \setlength{\tabcolsep}{1.5mm}
    \begin{tabular}{c|c|l|l|l}
    \toprule[1pt]
    Method                       & Partition & MAE$\downarrow$ & SA-SOR$\uparrow$ & SOR$\uparrow$    \\
    \hline
    \multirow{2}{*}{SOR \cite{fang2021salient}}         &           & 0.083  & 0.637  & 0.836 \\
                                 & $\surd$   & $0.078_{\underline{-0.005}}$  & $0.657_{\underline{+0.020}}$  & $0.859_{\underline{+0.023}}$  \\
    \multirow{2}{*}{SOTR \cite{guo2021sotr}}        &           & 0.092  & 0.682  & 0.867  \\
                                 & $\surd$   & $0.081_{\underline{-0.011}}$ & $0.711_{\underline{+0.029}}$  & $0.893_{\underline{+0.026}}$   \\
    \multirow{2}{*}{M2F \cite{cheng2022masked}} &           & 0.085  & 0.694  & 0.862  \\
                                 & $\surd$   & $0.077_{\underline{-0.008}}$ & $0.721_{\underline{+0.027}}$ & $0.880_{\underline{+0.018}}$ \\
    \multirow{2}{*}{SOLOv2 \cite{wang2020solov2}}      &           & 0.096  & 0.676  & 0.848  \\
                                 & $\surd$   & $0.080_{\underline{-0.016}}$  & $0.713_{\underline{+0.037}}$  & $0.871_{\underline{+0.023}}$  \\
    \bottomrule[1pt]
    \end{tabular}
\end{table}
We propose a ranking by partition paradigm for the task of saliency ranking that can mitigate the effects of label ambiguity. The ranking by partition paradigm, in principle, can be applied to other models and enhance their performance on saliency ranking. We compare different models using ranking by partition versus the traditional ranking by sorting. As shown in \tabref{table:Ablation of partition}, the ranking by partition paradigm improves the performance of saliency ranking not only on CNN-based models \cite{wang2020solov2}, but also on CNN-transformer-hybrid-based models \cite{guo2021sotr, fang2021salient} and mask-classification-based models \cite{cheng2022masked}, demonstrating its generalizability.

\section{CONCLUSIONS}
In this paper, we propose a ranking by partition paradigm to alleviate saliency ambiguity in the saliency ranking task. The ranking by partition paradigm can be applied to other related models to boost saliency ranking. We also proposed a dense pyramid transformer (DPT) to facilitate cross-scale global feature interaction for saliency ranking. By leveraging the proposed ranking by partition and DPT, our partitioned saliency ranking (PSR) outperforms state-of-the-art methods.

\section{Acknowledgments}
This work was supported by the National Natural Science Foundation of China Grant 61902139.

\newpage
\bibliographystyle{ACM-Reference-Format}
\bibliography{main}


\begin{thebibliography}{51}


\ifx \showCODEN    \undefined \def \showCODEN     #1{\unskip}     \fi
\ifx \showDOI      \undefined \def \showDOI       #1{#1}\fi
\ifx \showISBNx    \undefined \def \showISBNx     #1{\unskip}     \fi
\ifx \showISBNxiii \undefined \def \showISBNxiii  #1{\unskip}     \fi
\ifx \showISSN     \undefined \def \showISSN      #1{\unskip}     \fi
\ifx \showLCCN     \undefined \def \showLCCN      #1{\unskip}     \fi
\ifx \shownote     \undefined \def \shownote      #1{#1}          \fi
\ifx \showarticletitle \undefined \def \showarticletitle #1{#1}   \fi
\ifx \showURL      \undefined \def \showURL       {\relax}        \fi
\providecommand\bibfield[2]{#2}
\providecommand\bibinfo[2]{#2}
\providecommand\natexlab[1]{#1}
\providecommand\showeprint[2][]{arXiv:#2}

\bibitem[Cai and Vasconcelos(2018)]%
        {cai2018cascade}
\bibfield{author}{\bibinfo{person}{Zhaowei Cai} {and} \bibinfo{person}{Nuno
  Vasconcelos}.} \bibinfo{year}{2018}\natexlab{}.
\newblock \showarticletitle{Cascade r-cnn: Delving into high quality object
  detection}. In \bibinfo{booktitle}{\emph{Proceedings of the IEEE conference
  on computer vision and pattern recognition}}. \bibinfo{publisher}{IEEE},
  \bibinfo{address}{Salt Lake City, UT, USA}, \bibinfo{pages}{6154--6162}.
\newblock


\bibitem[Carion et~al\mbox{.}(2020)]%
        {carion2020end}
\bibfield{author}{\bibinfo{person}{Nicolas Carion}, \bibinfo{person}{Francisco
  Massa}, \bibinfo{person}{Gabriel Synnaeve}, \bibinfo{person}{Nicolas
  Usunier}, \bibinfo{person}{Alexander Kirillov}, {and} \bibinfo{person}{Sergey
  Zagoruyko}.} \bibinfo{year}{2020}\natexlab{}.
\newblock \showarticletitle{End-to-end object detection with transformers}. In
  \bibinfo{booktitle}{\emph{Computer Vision--ECCV 2020: 16th European
  Conference, Glasgow, UK, August 23--28, 2020, Proceedings, Part I 16}}.
  \bibinfo{publisher}{Springer}, \bibinfo{pages}{213--229}.
\newblock


\bibitem[Chen et~al\mbox{.}(2022)]%
        {chen2022saliency}
\bibfield{author}{\bibinfo{person}{Cuiqun Chen}, \bibinfo{person}{Mang Ye},
  \bibinfo{person}{Meibin Qi}, \bibinfo{person}{Jingjing Wu},
  \bibinfo{person}{Yimin Liu}, {and} \bibinfo{person}{Jianguo Jiang}.}
  \bibinfo{year}{2022}\natexlab{}.
\newblock \showarticletitle{Saliency and granularity: Discovering temporal
  coherence for video-based person re-identification}.
\newblock \bibinfo{journal}{\emph{IEEE Transactions on Circuits and Systems for
  Video Technology}} \bibinfo{volume}{32}, \bibinfo{number}{9}
  (\bibinfo{year}{2022}), \bibinfo{pages}{6100--6112}.
\newblock


\bibitem[Cheng et~al\mbox{.}(2022a)]%
        {cheng2022masked}
\bibfield{author}{\bibinfo{person}{Bowen Cheng}, \bibinfo{person}{Ishan Misra},
  \bibinfo{person}{Alexander~G Schwing}, \bibinfo{person}{Alexander Kirillov},
  {and} \bibinfo{person}{Rohit Girdhar}.} \bibinfo{year}{2022}\natexlab{a}.
\newblock \showarticletitle{Masked-attention mask transformer for universal
  image segmentation}. In \bibinfo{booktitle}{\emph{Proceedings of the IEEE/CVF
  Conference on Computer Vision and Pattern Recognition}}.
  \bibinfo{publisher}{{IEEE}}, \bibinfo{address}{New Orleans, LA, USA},
  \bibinfo{pages}{1290--1299}.
\newblock


\bibitem[Cheng et~al\mbox{.}(2021)]%
        {cheng2021per}
\bibfield{author}{\bibinfo{person}{Bowen Cheng}, \bibinfo{person}{Alex
  Schwing}, {and} \bibinfo{person}{Alexander Kirillov}.}
  \bibinfo{year}{2021}\natexlab{}.
\newblock \showarticletitle{Per-pixel classification is not all you need for
  semantic segmentation}.
\newblock \bibinfo{journal}{\emph{Advances in Neural Information Processing
  Systems}}  \bibinfo{volume}{34} (\bibinfo{year}{2021}),
  \bibinfo{pages}{17864--17875}.
\newblock


\bibitem[Cheng et~al\mbox{.}(2014)]%
        {cheng2014global}
\bibfield{author}{\bibinfo{person}{Ming-Ming Cheng}, \bibinfo{person}{Niloy~J
  Mitra}, \bibinfo{person}{Xiaolei Huang}, \bibinfo{person}{Philip~HS Torr},
  {and} \bibinfo{person}{Shi-Min Hu}.} \bibinfo{year}{2014}\natexlab{}.
\newblock \showarticletitle{Global contrast based salient region detection}.
\newblock \bibinfo{journal}{\emph{IEEE transactions on pattern analysis and
  machine intelligence}} \bibinfo{volume}{37}, \bibinfo{number}{3}
  (\bibinfo{year}{2014}), \bibinfo{pages}{569--582}.
\newblock


\bibitem[Cheng et~al\mbox{.}(2022b)]%
        {cheng2022sparse}
\bibfield{author}{\bibinfo{person}{Tianheng Cheng}, \bibinfo{person}{Xinggang
  Wang}, \bibinfo{person}{Shaoyu Chen}, \bibinfo{person}{Wenqiang Zhang},
  \bibinfo{person}{Qian Zhang}, \bibinfo{person}{Chang Huang},
  \bibinfo{person}{Zhaoxiang Zhang}, {and} \bibinfo{person}{Wenyu Liu}.}
  \bibinfo{year}{2022}\natexlab{b}.
\newblock \showarticletitle{Sparse instance activation for real-time instance
  segmentation}. In \bibinfo{booktitle}{\emph{Proceedings of the IEEE/CVF
  Conference on Computer Vision and Pattern Recognition}}.
  \bibinfo{pages}{4433--4442}.
\newblock


\bibitem[Fan et~al\mbox{.}(2019b)]%
        {fan2019understanding}
\bibfield{author}{\bibinfo{person}{Lifeng Fan}, \bibinfo{person}{Wenguan Wang},
  \bibinfo{person}{Siyuan Huang}, \bibinfo{person}{Xinyu Tang}, {and}
  \bibinfo{person}{Song-Chun Zhu}.} \bibinfo{year}{2019}\natexlab{b}.
\newblock \showarticletitle{Understanding human gaze communication by
  spatio-temporal graph reasoning}. In \bibinfo{booktitle}{\emph{Proceedings of
  the IEEE/CVF International Conference on Computer Vision}}.
  \bibinfo{pages}{5724--5733}.
\newblock


\bibitem[Fan et~al\mbox{.}(2019a)]%
        {fan2019s4net}
\bibfield{author}{\bibinfo{person}{Ruochen Fan}, \bibinfo{person}{Ming-Ming
  Cheng}, \bibinfo{person}{Qibin Hou}, \bibinfo{person}{Tai-Jiang Mu},
  \bibinfo{person}{Jingdong Wang}, {and} \bibinfo{person}{Shi-Min Hu}.}
  \bibinfo{year}{2019}\natexlab{a}.
\newblock \showarticletitle{S4net: Single stage salient-instance segmentation}.
  In \bibinfo{booktitle}{\emph{Proceedings of the IEEE/CVF conference on
  computer vision and pattern recognition}}. \bibinfo{pages}{6103--6112}.
\newblock


\bibitem[Fang et~al\mbox{.}(2021b)]%
        {fang2021salient}
\bibfield{author}{\bibinfo{person}{Hao Fang}, \bibinfo{person}{Daoxin Zhang},
  \bibinfo{person}{Yi Zhang}, \bibinfo{person}{Minghao Chen},
  \bibinfo{person}{Jiawei Li}, \bibinfo{person}{Yao Hu}, \bibinfo{person}{Deng
  Cai}, {and} \bibinfo{person}{Xiaofei He}.} \bibinfo{year}{2021}\natexlab{b}.
\newblock \showarticletitle{Salient object ranking with position-preserved
  attention}. In \bibinfo{booktitle}{\emph{Proceedings of the IEEE/CVF
  International Conference on Computer Vision}}. \bibinfo{pages}{16331--16341}.
\newblock


\bibitem[Fang et~al\mbox{.}(2021a)]%
        {fang2021instances}
\bibfield{author}{\bibinfo{person}{Yuxin Fang}, \bibinfo{person}{Shusheng
  Yang}, \bibinfo{person}{Xinggang Wang}, \bibinfo{person}{Yu Li},
  \bibinfo{person}{Chen Fang}, \bibinfo{person}{Ying Shan},
  \bibinfo{person}{Bin Feng}, {and} \bibinfo{person}{Wenyu Liu}.}
  \bibinfo{year}{2021}\natexlab{a}.
\newblock \showarticletitle{Instances as queries}. In
  \bibinfo{booktitle}{\emph{Proceedings of the IEEE/CVF international
  conference on computer vision}}. \bibinfo{pages}{6910--6919}.
\newblock


\bibitem[Feng et~al\mbox{.}(2019)]%
        {feng2019attentive}
\bibfield{author}{\bibinfo{person}{Mengyang Feng}, \bibinfo{person}{Huchuan
  Lu}, {and} \bibinfo{person}{Errui Ding}.} \bibinfo{year}{2019}\natexlab{}.
\newblock \showarticletitle{Attentive feedback network for boundary-aware
  salient object detection}. In \bibinfo{booktitle}{\emph{Proceedings of the
  IEEE/CVF conference on computer vision and pattern recognition}}.
  \bibinfo{pages}{1623--1632}.
\newblock


\bibitem[Gao et~al\mbox{.}(2020)]%
        {gao2020highly}
\bibfield{author}{\bibinfo{person}{Shang-Hua Gao}, \bibinfo{person}{Yong-Qiang
  Tan}, \bibinfo{person}{Ming-Ming Cheng}, \bibinfo{person}{Chengze Lu},
  \bibinfo{person}{Yunpeng Chen}, {and} \bibinfo{person}{Shuicheng Yan}.}
  \bibinfo{year}{2020}\natexlab{}.
\newblock \showarticletitle{Highly efficient salient object detection with 100k
  parameters}. In \bibinfo{booktitle}{\emph{Computer Vision--ECCV 2020: 16th
  European Conference, Glasgow, UK, August 23--28, 2020, Proceedings, Part
  VI}}. Springer, \bibinfo{pages}{702--721}.
\newblock


\bibitem[Guo et~al\mbox{.}(2021)]%
        {guo2021sotr}
\bibfield{author}{\bibinfo{person}{Ruohao Guo}, \bibinfo{person}{Dantong Niu},
  \bibinfo{person}{Liao Qu}, {and} \bibinfo{person}{Zhenbo Li}.}
  \bibinfo{year}{2021}\natexlab{}.
\newblock \showarticletitle{Sotr: Segmenting objects with transformers}. In
  \bibinfo{booktitle}{\emph{Proceedings of the IEEE/CVF International
  Conference on Computer Vision}}. \bibinfo{pages}{7157--7166}.
\newblock


\bibitem[He et~al\mbox{.}(2017)]%
        {he2017mask}
\bibfield{author}{\bibinfo{person}{Kaiming He}, \bibinfo{person}{Georgia
  Gkioxari}, \bibinfo{person}{Piotr Doll{\'a}r}, {and} \bibinfo{person}{Ross
  Girshick}.} \bibinfo{year}{2017}\natexlab{}.
\newblock \showarticletitle{Mask r-cnn}. In
  \bibinfo{booktitle}{\emph{Proceedings of the IEEE international conference on
  computer vision}}. \bibinfo{pages}{2961--2969}.
\newblock


\bibitem[He et~al\mbox{.}(2016)]%
        {he2016deep}
\bibfield{author}{\bibinfo{person}{Kaiming He}, \bibinfo{person}{Xiangyu
  Zhang}, \bibinfo{person}{Shaoqing Ren}, {and} \bibinfo{person}{Jian Sun}.}
  \bibinfo{year}{2016}\natexlab{}.
\newblock \showarticletitle{Deep residual learning for image recognition}. In
  \bibinfo{booktitle}{\emph{Proceedings of the IEEE conference on computer
  vision and pattern recognition}}. \bibinfo{pages}{770--778}.
\newblock


\bibitem[Islam et~al\mbox{.}(2018)]%
        {islam2018revisiting}
\bibfield{author}{\bibinfo{person}{Md~Amirul Islam}, \bibinfo{person}{Mahmoud
  Kalash}, {and} \bibinfo{person}{Neil~DB Bruce}.}
  \bibinfo{year}{2018}\natexlab{}.
\newblock \showarticletitle{Revisiting salient object detection: Simultaneous
  detection, ranking, and subitizing of multiple salient objects}. In
  \bibinfo{booktitle}{\emph{Proceedings of the IEEE conference on computer
  vision and pattern recognition}}. \bibinfo{pages}{7142--7150}.
\newblock


\bibitem[Li et~al\mbox{.}(2017)]%
        {li2017instance}
\bibfield{author}{\bibinfo{person}{Guanbin Li}, \bibinfo{person}{Yuan Xie},
  \bibinfo{person}{Liang Lin}, {and} \bibinfo{person}{Yizhou Yu}.}
  \bibinfo{year}{2017}\natexlab{}.
\newblock \showarticletitle{Instance-level salient object segmentation}. In
  \bibinfo{booktitle}{\emph{Proceedings of the IEEE conference on computer
  vision and pattern recognition}}. \bibinfo{pages}{2386--2395}.
\newblock


\bibitem[Lin et~al\mbox{.}(2017a)]%
        {lin2017feature}
\bibfield{author}{\bibinfo{person}{Tsung-Yi Lin}, \bibinfo{person}{Piotr
  Doll{\'a}r}, \bibinfo{person}{Ross Girshick}, \bibinfo{person}{Kaiming He},
  \bibinfo{person}{Bharath Hariharan}, {and} \bibinfo{person}{Serge Belongie}.}
  \bibinfo{year}{2017}\natexlab{a}.
\newblock \showarticletitle{Feature pyramid networks for object detection}. In
  \bibinfo{booktitle}{\emph{Proceedings of the IEEE conference on computer
  vision and pattern recognition}}. \bibinfo{pages}{2117--2125}.
\newblock


\bibitem[Lin et~al\mbox{.}(2017b)]%
        {lin2017focal}
\bibfield{author}{\bibinfo{person}{Tsung-Yi Lin}, \bibinfo{person}{Priya
  Goyal}, \bibinfo{person}{Ross Girshick}, \bibinfo{person}{Kaiming He}, {and}
  \bibinfo{person}{Piotr Doll{\'a}r}.} \bibinfo{year}{2017}\natexlab{b}.
\newblock \showarticletitle{Focal loss for dense object detection}. In
  \bibinfo{booktitle}{\emph{Proceedings of the IEEE international conference on
  computer vision}}. \bibinfo{pages}{2980--2988}.
\newblock


\bibitem[Lin et~al\mbox{.}(2014)]%
        {lin2014microsoft}
\bibfield{author}{\bibinfo{person}{Tsung-Yi Lin}, \bibinfo{person}{Michael
  Maire}, \bibinfo{person}{Serge Belongie}, \bibinfo{person}{James Hays},
  \bibinfo{person}{Pietro Perona}, \bibinfo{person}{Deva Ramanan},
  \bibinfo{person}{Piotr Doll{\'a}r}, {and} \bibinfo{person}{C~Lawrence
  Zitnick}.} \bibinfo{year}{2014}\natexlab{}.
\newblock \showarticletitle{Microsoft coco: Common objects in context}. In
  \bibinfo{booktitle}{\emph{Computer Vision--ECCV 2014: 13th European
  Conference, Zurich, Switzerland, September 6-12, 2014, Proceedings, Part V
  13}}. Springer, \bibinfo{pages}{740--755}.
\newblock


\bibitem[Liu et~al\mbox{.}(2021a)]%
        {liu2021instance}
\bibfield{author}{\bibinfo{person}{Nian Liu}, \bibinfo{person}{Long Li},
  \bibinfo{person}{Wangbo Zhao}, \bibinfo{person}{Junwei Han}, {and}
  \bibinfo{person}{Ling Shao}.} \bibinfo{year}{2021}\natexlab{a}.
\newblock \showarticletitle{Instance-level relative saliency ranking with graph
  reasoning}.
\newblock \bibinfo{journal}{\emph{IEEE Transactions on Pattern Analysis and
  Machine Intelligence}} \bibinfo{volume}{44}, \bibinfo{number}{11}
  (\bibinfo{year}{2021}), \bibinfo{pages}{8321--8337}.
\newblock


\bibitem[Liu et~al\mbox{.}(2021c)]%
        {liu2021scg}
\bibfield{author}{\bibinfo{person}{Nian Liu}, \bibinfo{person}{Wangbo Zhao},
  \bibinfo{person}{Ling Shao}, {and} \bibinfo{person}{Junwei Han}.}
  \bibinfo{year}{2021}\natexlab{c}.
\newblock \showarticletitle{SCG: Saliency and contour guided salient instance
  segmentation}.
\newblock \bibinfo{journal}{\emph{IEEE Transactions on Image Processing}}
  \bibinfo{volume}{30} (\bibinfo{year}{2021}), \bibinfo{pages}{5862--5874}.
\newblock


\bibitem[Liu et~al\mbox{.}(2010)]%
        {liu2010learning}
\bibfield{author}{\bibinfo{person}{Tie Liu}, \bibinfo{person}{Zejian Yuan},
  \bibinfo{person}{Jian Sun}, \bibinfo{person}{Jingdong Wang},
  \bibinfo{person}{Nanning Zheng}, \bibinfo{person}{Xiaoou Tang}, {and}
  \bibinfo{person}{Heung-Yeung Shum}.} \bibinfo{year}{2010}\natexlab{}.
\newblock \showarticletitle{Learning to detect a salient object}.
\newblock \bibinfo{journal}{\emph{IEEE Transactions on Pattern analysis and
  machine intelligence}} \bibinfo{volume}{33}, \bibinfo{number}{2}
  (\bibinfo{year}{2010}), \bibinfo{pages}{353--367}.
\newblock


\bibitem[Liu et~al\mbox{.}(2021b)]%
        {liu2021swin}
\bibfield{author}{\bibinfo{person}{Ze Liu}, \bibinfo{person}{Yutong Lin},
  \bibinfo{person}{Yue Cao}, \bibinfo{person}{Han Hu}, \bibinfo{person}{Yixuan
  Wei}, \bibinfo{person}{Zheng Zhang}, \bibinfo{person}{Stephen Lin}, {and}
  \bibinfo{person}{Baining Guo}.} \bibinfo{year}{2021}\natexlab{b}.
\newblock \showarticletitle{Swin transformer: Hierarchical vision transformer
  using shifted windows}. In \bibinfo{booktitle}{\emph{Proceedings of the
  IEEE/CVF international conference on computer vision}}.
  \bibinfo{pages}{10012--10022}.
\newblock


\bibitem[Long et~al\mbox{.}(2015)]%
        {long2015fully}
\bibfield{author}{\bibinfo{person}{Jonathan Long}, \bibinfo{person}{Evan
  Shelhamer}, {and} \bibinfo{person}{Trevor Darrell}.}
  \bibinfo{year}{2015}\natexlab{}.
\newblock \showarticletitle{Fully convolutional networks for semantic
  segmentation}. In \bibinfo{booktitle}{\emph{Proceedings of the IEEE
  conference on computer vision and pattern recognition}}.
  \bibinfo{pages}{3431--3440}.
\newblock


\bibitem[Ma et~al\mbox{.}(2021)]%
        {ma2021implicit}
\bibfield{author}{\bibinfo{person}{Lufan Ma}, \bibinfo{person}{Tiancai Wang},
  \bibinfo{person}{Bin Dong}, \bibinfo{person}{Jiangpeng Yan},
  \bibinfo{person}{Xiu Li}, {and} \bibinfo{person}{Xiangyu Zhang}.}
  \bibinfo{year}{2021}\natexlab{}.
\newblock \showarticletitle{Implicit feature refinement for instance
  segmentation}. In \bibinfo{booktitle}{\emph{Proceedings of the 29th ACM
  International Conference on Multimedia}}. \bibinfo{pages}{3088--3096}.
\newblock


\bibitem[Park and Kim(2022)]%
        {park2022vision}
\bibfield{author}{\bibinfo{person}{Namuk Park} {and} \bibinfo{person}{Songkuk
  Kim}.} \bibinfo{year}{2022}\natexlab{}.
\newblock \showarticletitle{How do vision transformers work?}
\newblock \bibinfo{journal}{\emph{arXiv preprint arXiv:2202.06709}}
  (\bibinfo{year}{2022}).
\newblock


\bibitem[Pei et~al\mbox{.}(2022a)]%
        {pei2022osformer}
\bibfield{author}{\bibinfo{person}{Jialun Pei}, \bibinfo{person}{Tianyang
  Cheng}, \bibinfo{person}{Deng-Ping Fan}, \bibinfo{person}{He Tang},
  \bibinfo{person}{Chuanbo Chen}, {and} \bibinfo{person}{Luc Van~Gool}.}
  \bibinfo{year}{2022}\natexlab{a}.
\newblock \showarticletitle{Osformer: One-stage camouflaged instance
  segmentation with transformers}. In \bibinfo{booktitle}{\emph{Computer
  Vision--ECCV 2022: 17th European Conference, Tel Aviv, Israel, October
  23--27, 2022, Proceedings, Part XVIII}}. Springer, \bibinfo{pages}{19--37}.
\newblock


\bibitem[Pei et~al\mbox{.}(2022b)]%
        {pei2022transformer}
\bibfield{author}{\bibinfo{person}{Jialun Pei}, \bibinfo{person}{Tianyang
  Cheng}, \bibinfo{person}{He Tang}, {and} \bibinfo{person}{Chuanbo Chen}.}
  \bibinfo{year}{2022}\natexlab{b}.
\newblock \showarticletitle{Transformer-based efficient salient instance
  segmentation networks with orientative query}.
\newblock \bibinfo{journal}{\emph{IEEE Transactions on Multimedia}}
  (\bibinfo{year}{2022}).
\newblock


\bibitem[Pei et~al\mbox{.}(2022c)]%
        {pei2022salient}
\bibfield{author}{\bibinfo{person}{Jialun Pei}, \bibinfo{person}{He Tang},
  \bibinfo{person}{Wanru Wang}, \bibinfo{person}{Tianyang Cheng}, {and}
  \bibinfo{person}{Chuanbo Chen}.} \bibinfo{year}{2022}\natexlab{c}.
\newblock \showarticletitle{Salient instance segmentation with region and
  box-level annotations}.
\newblock \bibinfo{journal}{\emph{Neurocomputing}}  \bibinfo{volume}{507}
  (\bibinfo{year}{2022}), \bibinfo{pages}{332--344}.
\newblock


\bibitem[Qin et~al\mbox{.}(2021)]%
        {qin2021learning}
\bibfield{author}{\bibinfo{person}{Zheyun Qin}, \bibinfo{person}{Xiankai Lu},
  \bibinfo{person}{Xiushan Nie}, \bibinfo{person}{Xiantong Zhen}, {and}
  \bibinfo{person}{Yilong Yin}.} \bibinfo{year}{2021}\natexlab{}.
\newblock \showarticletitle{Learning hierarchical embedding for video instance
  segmentation}. In \bibinfo{booktitle}{\emph{Proceedings of the 29th ACM
  International Conference on Multimedia}}. \bibinfo{pages}{1884--1892}.
\newblock


\bibitem[Ren et~al\mbox{.}(2015)]%
        {ren2015faster}
\bibfield{author}{\bibinfo{person}{Shaoqing Ren}, \bibinfo{person}{Kaiming He},
  \bibinfo{person}{Ross Girshick}, {and} \bibinfo{person}{Jian Sun}.}
  \bibinfo{year}{2015}\natexlab{}.
\newblock \showarticletitle{Faster r-cnn: Towards real-time object detection
  with region proposal networks}.
\newblock \bibinfo{journal}{\emph{Advances in neural information processing
  systems}}  \bibinfo{volume}{28} (\bibinfo{year}{2015}).
\newblock


\bibitem[RichardWebster et~al\mbox{.}(2022)]%
        {richardwebster2022doppelganger}
\bibfield{author}{\bibinfo{person}{Brandon RichardWebster},
  \bibinfo{person}{Brian Hu}, \bibinfo{person}{Keith Fieldhouse}, {and}
  \bibinfo{person}{Anthony Hoogs}.} \bibinfo{year}{2022}\natexlab{}.
\newblock \showarticletitle{Doppelganger Saliency: Towards More Ethical Person
  Re-Identification}. In \bibinfo{booktitle}{\emph{Proceedings of the IEEE/CVF
  Conference on Computer Vision and Pattern Recognition}}.
  \bibinfo{pages}{2847--2857}.
\newblock


\bibitem[Seth and Bayne(2022)]%
        {seth2022theories}
\bibfield{author}{\bibinfo{person}{Anil~K Seth} {and} \bibinfo{person}{Tim
  Bayne}.} \bibinfo{year}{2022}\natexlab{}.
\newblock \showarticletitle{Theories of consciousness}.
\newblock \bibinfo{journal}{\emph{Nature Reviews Neuroscience}}
  \bibinfo{volume}{23}, \bibinfo{number}{7} (\bibinfo{year}{2022}),
  \bibinfo{pages}{439--452}.
\newblock


\bibitem[Siris et~al\mbox{.}(2020)]%
        {siris2020inferring}
\bibfield{author}{\bibinfo{person}{Avishek Siris}, \bibinfo{person}{Jianbo
  Jiao}, \bibinfo{person}{Gary~KL Tam}, \bibinfo{person}{Xianghua Xie}, {and}
  \bibinfo{person}{Rynson~WH Lau}.} \bibinfo{year}{2020}\natexlab{}.
\newblock \showarticletitle{Inferring attention shift ranks of objects for
  image saliency}. In \bibinfo{booktitle}{\emph{Proceedings of the IEEE/CVF
  Conference on Computer Vision and Pattern Recognition}}.
  \bibinfo{pages}{12133--12143}.
\newblock


\bibitem[Siris et~al\mbox{.}(2021)]%
        {siris2021scene}
\bibfield{author}{\bibinfo{person}{Avishek Siris}, \bibinfo{person}{Jianbo
  Jiao}, \bibinfo{person}{Gary~KL Tam}, \bibinfo{person}{Xianghua Xie}, {and}
  \bibinfo{person}{Rynson~WH Lau}.} \bibinfo{year}{2021}\natexlab{}.
\newblock \showarticletitle{Scene context-aware salient object detection}. In
  \bibinfo{booktitle}{\emph{Proceedings of the IEEE/CVF International
  Conference on Computer Vision}}. \bibinfo{pages}{4156--4166}.
\newblock


\bibitem[Tian et~al\mbox{.}(2022)]%
        {tian2022bi}
\bibfield{author}{\bibinfo{person}{Xin Tian}, \bibinfo{person}{Ke Xu},
  \bibinfo{person}{Xin Yang}, \bibinfo{person}{Lin Du}, \bibinfo{person}{Baocai
  Yin}, {and} \bibinfo{person}{Rynson~WH Lau}.}
  \bibinfo{year}{2022}\natexlab{}.
\newblock \showarticletitle{Bi-directional object-context prioritization
  learning for saliency ranking}. In \bibinfo{booktitle}{\emph{Proceedings of
  the IEEE/CVF Conference on Computer Vision and Pattern Recognition}}.
  \bibinfo{pages}{5882--5891}.
\newblock


\bibitem[Tian et~al\mbox{.}(2020)]%
        {tian2020weakly}
\bibfield{author}{\bibinfo{person}{Xin Tian}, \bibinfo{person}{Ke Xu},
  \bibinfo{person}{Xin Yang}, \bibinfo{person}{Baocai Yin}, {and}
  \bibinfo{person}{Rynson~WH Lau}.} \bibinfo{year}{2020}\natexlab{}.
\newblock \showarticletitle{Weakly-supervised salient instance detection}.
\newblock \bibinfo{journal}{\emph{arXiv preprint arXiv:2009.13898}}
  (\bibinfo{year}{2020}).
\newblock


\bibitem[Wang et~al\mbox{.}(2020a)]%
        {wang2020solo}
\bibfield{author}{\bibinfo{person}{Xinlong Wang}, \bibinfo{person}{Tao Kong},
  \bibinfo{person}{Chunhua Shen}, \bibinfo{person}{Yuning Jiang}, {and}
  \bibinfo{person}{Lei Li}.} \bibinfo{year}{2020}\natexlab{a}.
\newblock \showarticletitle{Solo: Segmenting objects by locations}. In
  \bibinfo{booktitle}{\emph{Computer Vision--ECCV 2020: 16th European
  Conference, Glasgow, UK, August 23--28, 2020, Proceedings, Part XVIII 16}}.
  Springer, \bibinfo{pages}{649--665}.
\newblock


\bibitem[Wang et~al\mbox{.}(2020b)]%
        {wang2020solov2}
\bibfield{author}{\bibinfo{person}{Xinlong Wang}, \bibinfo{person}{Rufeng
  Zhang}, \bibinfo{person}{Tao Kong}, \bibinfo{person}{Lei Li}, {and}
  \bibinfo{person}{Chunhua Shen}.} \bibinfo{year}{2020}\natexlab{b}.
\newblock \showarticletitle{Solov2: Dynamic and fast instance segmentation}.
\newblock \bibinfo{journal}{\emph{Advances in Neural information processing
  systems}}  \bibinfo{volume}{33} (\bibinfo{year}{2020}),
  \bibinfo{pages}{17721--17732}.
\newblock


\bibitem[Wang et~al\mbox{.}(2019)]%
        {wang2019ranking}
\bibfield{author}{\bibinfo{person}{Zheng Wang}, \bibinfo{person}{Xinyu Yan},
  \bibinfo{person}{Yahong Han}, {and} \bibinfo{person}{Meijun Sun}.}
  \bibinfo{year}{2019}\natexlab{}.
\newblock \showarticletitle{Ranking video salient object detection}. In
  \bibinfo{booktitle}{\emph{Proceedings of the 27th ACM International
  Conference on Multimedia}}. \bibinfo{pages}{873--881}.
\newblock


\bibitem[Wei et~al\mbox{.}(2020)]%
        {wei2020label}
\bibfield{author}{\bibinfo{person}{Jun Wei}, \bibinfo{person}{Shuhui Wang},
  \bibinfo{person}{Zhe Wu}, \bibinfo{person}{Chi Su}, \bibinfo{person}{Qingming
  Huang}, {and} \bibinfo{person}{Qi Tian}.} \bibinfo{year}{2020}\natexlab{}.
\newblock \showarticletitle{Label decoupling framework for salient object
  detection}. In \bibinfo{booktitle}{\emph{Proceedings of the IEEE/CVF
  conference on computer vision and pattern recognition}}.
  \bibinfo{pages}{13025--13034}.
\newblock


\bibitem[Wu and He(2018)]%
        {wu2018group}
\bibfield{author}{\bibinfo{person}{Yuxin Wu} {and} \bibinfo{person}{Kaiming
  He}.} \bibinfo{year}{2018}\natexlab{}.
\newblock \showarticletitle{Group normalization}. In
  \bibinfo{booktitle}{\emph{Proceedings of the European conference on computer
  vision (ECCV)}}. \bibinfo{pages}{3--19}.
\newblock


\bibitem[Wu et~al\mbox{.}(2021)]%
        {wu2021regularized}
\bibfield{author}{\bibinfo{person}{Yu-Huan Wu}, \bibinfo{person}{Yun Liu},
  \bibinfo{person}{Le Zhang}, \bibinfo{person}{Wang Gao}, {and}
  \bibinfo{person}{Ming-Ming Cheng}.} \bibinfo{year}{2021}\natexlab{}.
\newblock \showarticletitle{Regularized densely-connected pyramid network for
  salient instance segmentation}.
\newblock \bibinfo{journal}{\emph{IEEE Transactions on Image Processing}}
  \bibinfo{volume}{30} (\bibinfo{year}{2021}), \bibinfo{pages}{3897--3907}.
\newblock


\bibitem[Yang et~al\mbox{.}(2013)]%
        {yang2013saliency}
\bibfield{author}{\bibinfo{person}{Chuan Yang}, \bibinfo{person}{Lihe Zhang},
  \bibinfo{person}{Huchuan Lu}, \bibinfo{person}{Xiang Ruan}, {and}
  \bibinfo{person}{Ming-Hsuan Yang}.} \bibinfo{year}{2013}\natexlab{}.
\newblock \showarticletitle{Saliency detection via graph-based manifold
  ranking}. In \bibinfo{booktitle}{\emph{Proceedings of the IEEE conference on
  computer vision and pattern recognition}}. \bibinfo{pages}{3166--3173}.
\newblock


\bibitem[Zhang et~al\mbox{.}(2021)]%
        {zhang2021auto}
\bibfield{author}{\bibinfo{person}{Miao Zhang}, \bibinfo{person}{Tingwei Liu},
  \bibinfo{person}{Yongri Piao}, \bibinfo{person}{Shunyu Yao}, {and}
  \bibinfo{person}{Huchuan Lu}.} \bibinfo{year}{2021}\natexlab{}.
\newblock \showarticletitle{Auto-msfnet: Search multi-scale fusion network for
  salient object detection}. In \bibinfo{booktitle}{\emph{Proceedings of the
  29th ACM international conference on multimedia}}. \bibinfo{pages}{667--676}.
\newblock


\bibitem[Zhao et~al\mbox{.}(2019)]%
        {zhao2019egnet}
\bibfield{author}{\bibinfo{person}{Jia-Xing Zhao}, \bibinfo{person}{Jiang-Jiang
  Liu}, \bibinfo{person}{Deng-Ping Fan}, \bibinfo{person}{Yang Cao},
  \bibinfo{person}{Jufeng Yang}, {and} \bibinfo{person}{Ming-Ming Cheng}.}
  \bibinfo{year}{2019}\natexlab{}.
\newblock \showarticletitle{EGNet: Edge guidance network for salient object
  detection}. In \bibinfo{booktitle}{\emph{Proceedings of the IEEE/CVF
  international conference on computer vision}}. \bibinfo{pages}{8779--8788}.
\newblock


\bibitem[Zhao et~al\mbox{.}(2021)]%
        {zhao2021complementary}
\bibfield{author}{\bibinfo{person}{Zhirui Zhao}, \bibinfo{person}{Changqun
  Xia}, \bibinfo{person}{Chenxi Xie}, {and} \bibinfo{person}{Jia Li}.}
  \bibinfo{year}{2021}\natexlab{}.
\newblock \showarticletitle{Complementary trilateral decoder for fast and
  accurate salient object detection}. In \bibinfo{booktitle}{\emph{Proceedings
  of the 29th acm international conference on multimedia}}.
  \bibinfo{pages}{4967--4975}.
\newblock


\bibitem[Zhu et~al\mbox{.}(2021)]%
        {zhu2021horizontal}
\bibfield{author}{\bibinfo{person}{Tun Zhu}, \bibinfo{person}{Daoxin Zhang},
  \bibinfo{person}{Yao Hu}, \bibinfo{person}{Tianran Wang},
  \bibinfo{person}{Xiaolong Jiang}, \bibinfo{person}{Jianke Zhu}, {and}
  \bibinfo{person}{Jiawei Li}.} \bibinfo{year}{2021}\natexlab{}.
\newblock \showarticletitle{Horizontal-to-vertical video conversion}.
\newblock \bibinfo{journal}{\emph{IEEE Transactions on Multimedia}}
  \bibinfo{volume}{24} (\bibinfo{year}{2021}), \bibinfo{pages}{3036--3048}.
\newblock


\bibitem[Zhu et~al\mbox{.}(2020)]%
        {zhu2020deformable}
\bibfield{author}{\bibinfo{person}{Xizhou Zhu}, \bibinfo{person}{Weijie Su},
  \bibinfo{person}{Lewei Lu}, \bibinfo{person}{Bin Li},
  \bibinfo{person}{Xiaogang Wang}, {and} \bibinfo{person}{Jifeng Dai}.}
  \bibinfo{year}{2020}\natexlab{}.
\newblock \showarticletitle{Deformable DETR: Deformable Transformers for
  End-to-End Object Detection}. In \bibinfo{booktitle}{\emph{International
  Conference on Learning Representations}}.
\newblock


\end{thebibliography}

\appendix

\end{document}